\documentclass{article}

\PassOptionsToPackage{numbers, compress}{natbib}

\usepackage[main, final]{neurips_2026}
\usepackage[T1]{fontenc}
\usepackage{lmodern}

\renewcommand{\ttdefault}{lmtt}
\usepackage{url}
\usepackage{booktabs}
\usepackage{amsfonts}
\usepackage{amsmath}
\usepackage{nicefrac}
\usepackage{microtype}
\usepackage{xcolor}
\usepackage{graphicx}
\usepackage{tabularx}
\usepackage{array}
\usepackage{natbib}
\usepackage{fancyvrb}
\usepackage{placeins}
\usepackage{wrapfig}
\usepackage{needspace}
\usepackage{tcolorbox}
\tcbuselibrary{skins,breakable}

\definecolor{purple}{RGB}{111,36,174}
\newtcolorbox{keybox}[1]{
  enhanced,
  breakable,
  colback=purple!4,
  colframe=purple!4,
  boxrule=0pt,
  borderline west={3pt}{0pt}{purple},
  sharp corners,
  left=4mm,
  right=3mm,
  top=2mm,
  bottom=2mm,
  fontupper=\normalfont,
  before upper={\textbf{#1}\par\smallskip},
}

\newcommand{\promptboxcaption}[1]{%
\par\noindent\textbf{#1}\par\vspace{0.35em}%
}

\DefineVerbatimEnvironment{PromptBox}{Verbatim}{
  fontsize=\scriptsize,
  frame=single,
  framesep=6pt
}

\usepackage[pagebackref,breaklinks,colorlinks,citecolor=blue,linkcolor=red]{hyperref}
\usepackage{cleveref}

\title{How Do Agentic LLMs Decide to Call Tools?\\A Tool-Call Vector Shaped by Suppression}

\author{%
  \textbf{Xijie Gong}$^{1,2,*}$ \quad \textbf{Tingxu Han}$^{1,3,*,\dagger}$ \quad \textbf{Jiahao Zhang}$^{1}$ \quad \textbf{Wei Song}$^{4}$ \\
  \textbf{Ziqi Ding}$^{5}$ \quad \textbf{Hanqi Yan}$^{6}$ \quad \textbf{Youcheng Sun}$^{1}$ \quad \textbf{Lijie Hu}$^{1,\ddagger}$ \\[6pt]
  \small
  $^{1}$Mohamed bin Zayed University of Artificial Intelligence \\
  $^{2}$University of Electronic Science and Technology of China \\
  $^{3}$Nanjing University \\
  $^{4}$Griffith University \\
  $^{5}$University of New South Wales \\
  $^{6}$King's College London \\
}
\makeatletter
\def\@thanks{%
  \footnotetext[0]{$^{*}$Equal contribution \quad $^{\dagger}$Project lead \quad $^{\ddagger}$Corresponding author. Lijie Hu <lijie.hu@mbzuai.ac.ae>}%
}
\makeatother

\begin{document}

\maketitle

\begin{abstract}
Tool calling, invoking external tools on demand, is central to agentic LLMs, yet the mechanism that decides whether a model calls a tool or responds directly remains poorly understood.
Agentic prompts are long and heavily scaffolded, combining role instructions, tool schemas, format templates, and the user's request across hundreds of tokens, creating a noisy, highly entangled context in which no single controllable variable for mechanistic analysis is obvious.
To obtain such a variable, we propose a method that converts complex agentic prompts into minimal contrastive pairs in which a single request verb determines the tool-call decision: replacing an execution-verb (e.g., \textit{write}) with an analysis-verb (e.g., \textit{discuss}) reliably flips the decision, suggesting it is mediated by a compact internal state.
We construct 500 such paired prompts across Python, Java, and C++ (300 for mechanistic analysis, 200 held out for evaluation).
We trace the decision to a vector, $\mu_\Delta$, that is both causally necessary and sufficient and generalizes beyond the discovery prompts to native multi-turn $\tau^2$-Bench trajectories and verb-free requests. Behavioral ablations show that the scaffold establishes a tool-call prior; Transcoder decomposition then reveals that analysis verbs suppress this prior through features signaling that tool use is unnecessary, whereas execution verbs largely leave it intact. Downstream scaffold-reading attention heads and MLP features read out the resulting state, and the same mechanism recurs across seven models from the Qwen, Mistral, and Granite families.
Our code is available at \url{https://github.com/XijieGo/MI4ToolCalling}.
\end{abstract}

\section{Introduction}

AI agents, from personal AI assistants to software engineering agents, are now widely deployed across industry and research. 
These agents are powered by agentic LLMs whose core capability is tool calling: invoking external tools on demand to act in the world.
Given an agentic prompt containing a tool-call prompt scaffold (role instructions, tool schemas, and format templates) alongside the user's request, the LLM can generate a structured tool call specifying the selected tool and its arguments, enabling it to search the web, execute code \cite{schick2023toolformer, yao2023react}. 
For any agentic prompt, the LLM faces the decision of whether to call a tool or respond directly.
We call this the \emph{call-or-no-call} decision. An incorrect decision causes the model to call tools unnecessarily or fail to call a tool when one is needed.

\begin{figure}[!t]
    \centering
    \includegraphics[width=\linewidth]{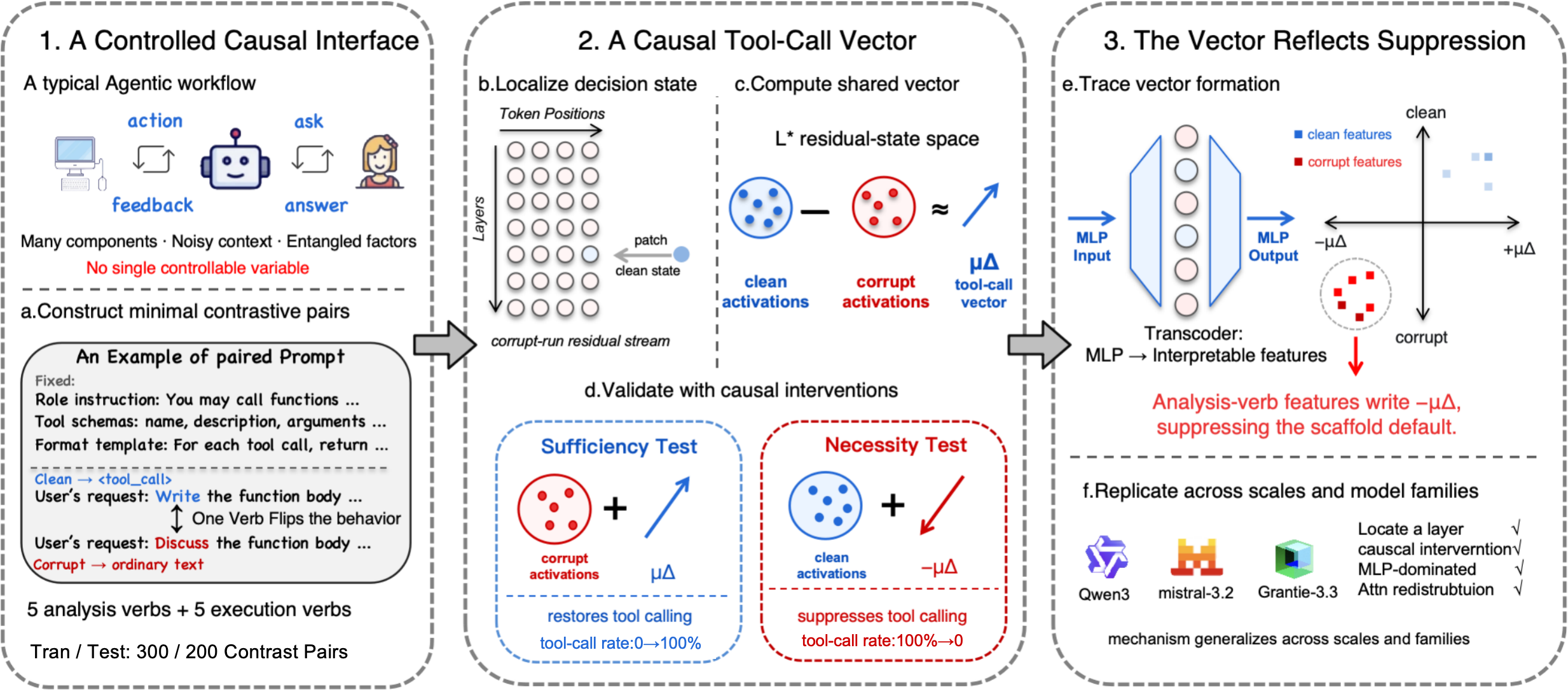}
    \caption{\textbf{Roadmap of the paper's mechanistic analysis.} The three numbered blocks summarize the stages of our analysis: (1) verb substitution turns long agentic prompts into a controlled causal interface; (2) the call-or-no-call decision passes through a tool-call vector, $\mu_\Delta$, that is causally sufficient and necessary; and (3) this vector reflects suppression of a scaffold-induced tool-call default and generalizes across the model families. The six lettered steps show how the evidence is built.}
    \label{fig:overview}
\end{figure}

Frontier LLMs with dedicated agentic training often signal this call-or-no-call decision through a special token.
In Qwen3, for instance, \texttt{<tool\_call>} appears as the first generated token precisely when the model decides to call a tool \citep{yang2025qwen3}.  
In this setting, the call-or-no-call decision reduces to a simple question: is \texttt{<tool\_call>} the model's top-1 prediction for the first generated token?
This makes the call-or-no-call decision tractable for mechanistic analysis: a single-token output prediction is the natural target for mechanistic tools such as activation patching and direct logit attribution.

Yet what drives this decision has not been studied mechanistically.
Existing work on tool calling has focused on whether LLMs select the right tools and how to improve their accuracy \citep{schick2023toolformer,patil2023gorilla, qin2023toolllm, huang2024metatool, patil2025bfcl}.
Prior mechanistic work has typically examined short prompts of 10--30 tokens, where a single variable can be varied while holding the rest fixed \citep{wang2022ioi, conmy2023acdc}. 
Agentic prompts are fundamentally different: the prompt scaffold (role instructions, tool schemas, and format templates) is packed alongside the user's request across several hundred tokens. 
With so many components that could each plausibly influence the decision, it is not clear which single element to vary to construct a reliable contrastive pair \citep{chen2025enhancing,qi2025agentif}.
This complexity makes it hard to isolate a single controllable variable for mechanistic analysis.

However, in code completion tasks, we find that the request verb serves as a controllable variable: replacing an execution verb with an analysis verb switches the first generated token in the retained pairs. Matched prompts differ only in the request verb, preserving the prompt scaffold, task body, and available tool. We construct 500 prompt pairs for each of the seven primary models, with 300 used for mechanistic analysis and 200 held out for evaluation. The coding tasks span Python, Java, and C++, drawing from MBPP, APPS, HumanEval, and CodeContests \citep{austin2021mbpp, hendrycks2021apps, chen2021codex, li2022alphacode}. By holding the surrounding prompt fixed while varying a single behaviorally decisive token, these pairs isolate a controlled interface for mechanistic analysis.

We study this mechanism primarily in Qwen3-8B and test whether it generalizes across model scales and families. Activation patching localizes the decisive residual-stream state to layer~24: patching it recovers \texttt{<tool\_call>} on 100\% of corrupt prompts. Averaging the activation difference at this layer gives the tool-call vector, $\mu_\Delta$. Adding $\mu_\Delta$ to analysis-verb prompts restores \texttt{<tool\_call>}; subtracting it from execution-verb prompts suppresses tool calling. Because these pairs are behaviorally filtered to exhibit the intended contrast, we test whether $\mu_\Delta$ generalizes beyond this construction. Furthermore, $\mu_\Delta$ induces and suppresses tool calls in web retrieval, SQL execution, and email dispatch with new tool schemas and request verbs. The same control extends to native decisions in multi-turn, multi-tool $\tau^2$-Bench trajectories across seven models, using model-specific $\mu_\Delta$ without re-estimation. Even for requests that imply a need for tools without containing the execution or analysis verbs used in our prompt pairs, subtracting $\mu_\Delta$ still suppresses tool calls. \textbf{The call-or-no-call decision passes through a compact tool-call vector shared across prompts.}

The natural hypothesis is that execution verbs actively generate a tool-call signal. Our evidence points to a different picture. The prompt scaffold establishes a tool-call prior: the role instructions, tool schemas, and format templates bias the model toward predicting \texttt{<tool\_call>} first on task-bearing requests rather than empty or unrelated turns. Analysis verbs activate a family of features that signal the tool is unnecessary; they write against this prior through layers~21--23. Execution verbs activate these features less strongly and largely preserve the prior. $\mu_\Delta$ is the inverse direction of the suppression signal that these features write into the residual stream. Downstream, scaffold-reading heads shift attention toward the tool-format template and MLP blocks support the opening of a tool-call sequence; both effects weaken when the suppression features are active. The mechanism replicates across the Qwen3-family and additional models, and structurally mirrors refusal \citep{arditi2024refusal}: a strong prior is present, and a compact suppressor overrides it. A scaffold-component ablation confirms this pattern behaviorally: the format template alone installs the near-ceiling call prior, and the tool schema makes it sensitive to wording.

To summarize, we make four contributions:
\textbf{(1)} A first-generated-token target and controlled prompt pairs provide a clean causal interface for studying the call-or-no-call decision in long, scaffolded prompts (\autoref{sec:interface}).
\textbf{(2)} A tool-call vector $\mu_\Delta$ at the layer-24 prediction position is causally necessary and sufficient for the call-or-no-call decision, and this control extends unchanged to real multi-turn agent trajectories and to requests with no cued verb at all (\autoref{sec:vector-state}, \autoref{sec:generalization}).
\textbf{(3)} $\mu_\Delta$ is the inverse of the suppression signal that features activated by analysis-verbs write against a scaffold-induced, request-sensitive call prior; a scaffold-component ablation clarifies this prior behaviorally (\autoref{sec:formation}).
\textbf{(4)} Downstream attention heads and MLP blocks read this vector into the first generated token, and the same mechanism extends across Qwen3 and additional models (\autoref{sec:readout}, \autoref{sec:crossscale}).

\section{One Verb Flips the Call-or-No-Call Decision}
\label{sec:interface}

\begin{wrapfigure}{r}{0.40\linewidth}
  \vspace{-10pt}
  \centering
  \includegraphics[width=\linewidth]{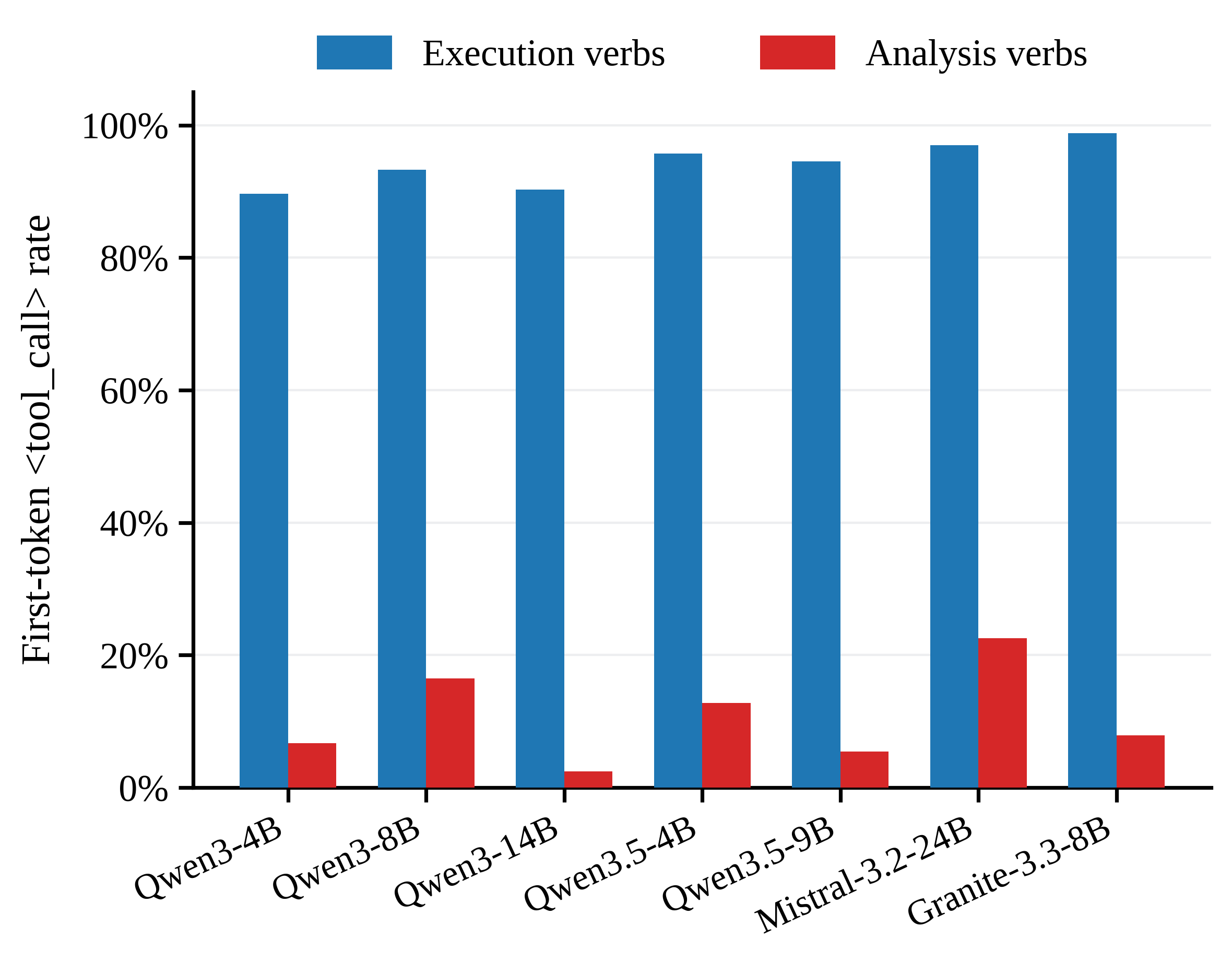}
  \caption{\textbf{Request verbs change tool-call rates across models.} Bars show first-token call rates on HumanEval before behavioral filtering, by request type.}
  \label{fig:interface}
  \vspace{-20pt}
\end{wrapfigure}

Mechanistic analysis requires a controlled change that produces a reliable behavioral contrast. Agentic prompts make such changes difficult to isolate. Role instructions, tool schemas, formatting rules, and user requests can all affect the decision. We seek a minimal change that exposes the call decision while preserving the task and scaffold.

We find a surprisingly simple control variable in code-completion tasks: a single request verb can reliably switch the model between calling a tool and responding directly. Asking the model to \textit{write} a function often triggers a tool call, while replacing \textit{write} with \textit{discuss} can instead produce a text response (Figure~\ref{fig:interface}). We therefore construct matched code-completion prompts that differ only in the request verb. Execution-verbs request an action on the code, and analysis-verbs request a textual response, while the task content, available tool, and scaffold remain fixed.

We study Qwen3-8B~\citep{yang2025qwen3} in non-thinking mode, where the first generated token exposes call initiation. A tool call begins with \texttt{<tool\_call>}, and a direct response begins with ordinary text. We use five execution verbs (add, build, ...) and five analysis verbs (discuss, explore, ...). Tasks come from MBPP, APPS, HumanEval, and CodeContests~\citep{austin2021mbpp,hendrycks2021apps,chen2021codex,li2022alphacode} and cover Python, Java, and C++. The Qwen3-8B dataset contains 500 pairs, with 300 for mechanistic analysis and 200 held out for evaluation. Appendix~\ref{appendix:dataset} summarizes the data sources, splits, and prompt format.

\Needspace{5\baselineskip}
\begin{keybox}{Key Setup: A One-Word Causal Interface}
Changing the request verb flips the first output between \texttt{<tool\_call>} and direct text. The matched prompts isolate the call decision while preserving the surrounding task and scaffold.
\end{keybox}

Following activation-patching terminology, we call execution prompts \emph{clean}, denoted $x^c$, and analysis prompts \emph{corrupt}, denoted $x^*$. Each prompt has the form
\begin{equation} x \;=\; \underbrace{[\,\text{role instructions},\;\text{tool schemas},\;\text{format templates}\,]}_{\text{prompt scaffold}} \;\; \underbrace{[\,v \;\|\; \text{task body}\,]}_{\text{user's request}} \;\longrightarrow\; y, \label{eq:prompt} \end{equation}
Here $v$ is the request verb, $y$ is the first generated token, and $y_{\mathrm{call}}=\texttt{<tool\_call>}$. The paired contrast is $y=y_{\mathrm{call}}$ on execution prompts and $y\neq y_{\mathrm{call}}$ on analysis prompts. Only $v$ changes within a pair. We write $h^{(l)}_p(x)\in\mathbb{R}^{d_{\mathrm{model}}}$ for the residual activation at layer $l$ and position $p$.

\FloatBarrier
\section{The Tool-Call Vector}
\label{sec:vector-state}

A shared residual vector controls the call-or-no-call decision in Qwen3-8B. We identify the vector in two steps. Activation patching localizes a \emph{tool-call state} at the final prompt position. The mean execution--analysis difference at that state gives the \emph{tool-call vector}.

\subsection{Localizing the Tool-Call State}
\label{sec:loc-where}

\begin{wrapfigure}{r}{0.38\linewidth}
  \vspace{-6pt}
  \centering
  \includegraphics[width=\linewidth]{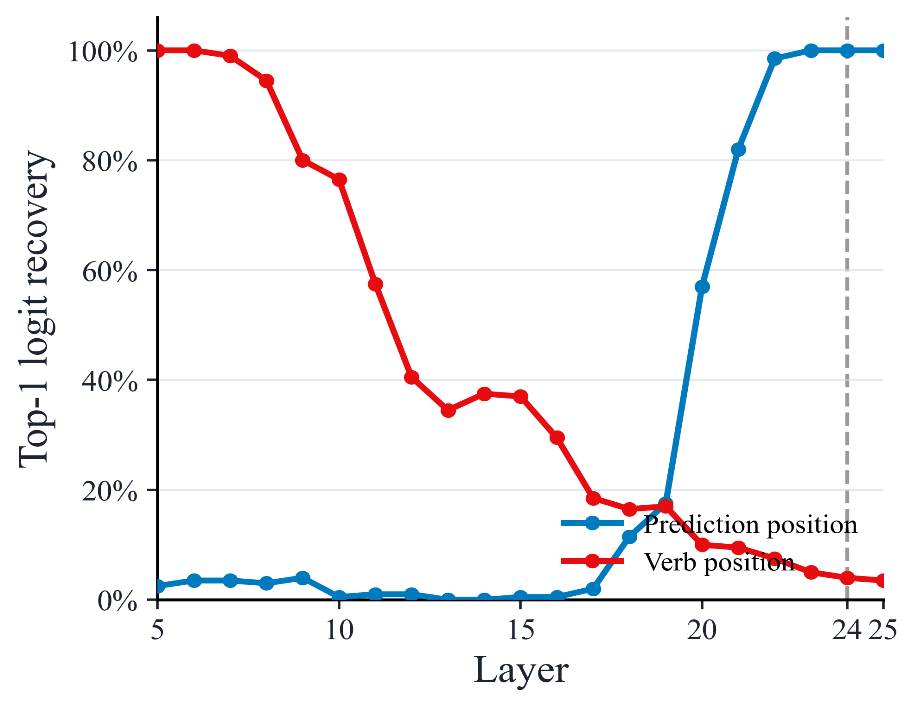}
  \caption{\textbf{The call signal shifts from the verb to the prediction position.} Layerwise patching finds the L24 residual state that determines when to call a tool or to give a text reply.}
  \label{fig:patching}
  \vspace{-6pt}
\end{wrapfigure}

Activation patching locates the request verb's effect on the call decision~\citep{meng2022locating,heimersheim2024patching}. For each held-out pair $(x_i^c,x_i^*)$, we patch the analysis state at layer $l$ and position $q$ with its execution counterpart. Other positions at that layer stay fixed; the rest of the model resumes from the patched state.
\begin{equation} \tilde{h}^{(l,q)}_j(x_i^*) = \begin{cases} h^{(l)}_q(x_i^c), & j=q,\\ h^{(l)}_j(x_i^*), & j\neq q. \end{cases} \label{eq:patch} \end{equation}
The recovery rate $r(l,q)$ is the fraction of patched prompts whose first-token top-1 prediction is $y_{\mathrm{call}}$. We sweep layers at the verb position and the \emph{prediction position}, the final prompt position that predicts the first output token. The patching effect shifts from the verb to the prediction position across layers. At L24, patching the prediction-position state recovers calls on 100\% of held-out analysis prompts. This state is where we add and remove the tool-call vector.

\subsection{Estimating the Tool-Call Vector}
\label{sec:shared-direction}

We estimate the tool-call vector by averaging the execution-minus-analysis activation difference over the 300 training pairs at the L24 prediction position $p$.
\begin{align} \Delta_i &= h^{(24)}_p(x_i^c) - h^{(24)}_p(x_i^*), \qquad \mu_\Delta = \operatorname*{mean}_i \Delta_i . \label{eq:vector-def} \end{align}
We then add $\mu_\Delta$ to analysis-prompt activations and subtract $\mu_\Delta$ from execution-prompt activations.
\begin{equation}
\text{Add:}\quad \tilde{h}^{(24)}_p(x_i^*) = h^{(24)}_p(x_i^*) + \mu_\Delta,
\qquad
\text{Remove:}\quad \tilde{h}^{(24)}_p(x_i^c) = h^{(24)}_p(x_i^c) - \mu_\Delta.
\label{eq:vector-interventions}
\end{equation}
Both interventions change a single residual activation before downstream computation resumes. We measure each intervention's effect relative to the original execution--analysis logit gap. Let $\bar z_{\mathrm{call}}^c$ and $\bar z_{\mathrm{call}}^*$ denote mean tool-call logits on held-out execution and analysis prompts. Superscripts $*+\mu$ and $c-\mu$ denote addition and removal. The normalized effects are
\begin{equation} \mathrm{Suff}(\mu_\Delta) = \frac{\bar{z}_{\mathrm{call}}^{*+\mu}-\bar{z}_{\mathrm{call}}^*} {\bar{z}_{\mathrm{call}}^c-\bar{z}_{\mathrm{call}}^*}, \qquad \mathrm{Necc}(\mu_\Delta) = \frac{\bar{z}_{\mathrm{call}}^c-\bar{z}_{\mathrm{call}}^{c-\mu}} {\bar{z}_{\mathrm{call}}^c-\bar{z}_{\mathrm{call}}^*}. \label{eq:suff-necc} \end{equation}
Addition and removal recover most of the logit gap in their respective directions (\autoref{tab:shared-vector}). The same vector therefore supports both tool-call induction and suppression at the prediction position. 

\begin{table}[h]
\vspace{-4pt}
\centering
\small
\caption{\textbf{The tool-call vector causally controls the call-or-no-call decision.} Adding $\mu_\Delta$ to analysis prompts induces tool calls, while removing it from execution prompts suppresses them.}
\label{tab:shared-vector}
\setlength{\tabcolsep}{3.5pt}
\begin{tabularx}{\linewidth}{>{\raggedright\arraybackslash}Xcccccccc}
\toprule
Intervention & Top-1 before & Top-1 after & $\Delta$ top-1 & Logit before & Logit after & $\Delta$ logit & Suff. & Necc. \\
\midrule
Add $\mu_\Delta$ & 0.00\% & 100.00\% & $\textcolor{red}{\uparrow}\,100.00$ & 25.11 & 32.50 & $\textcolor{red}{\uparrow}\,7.39$ & 1.03 & -- \\
Remove $\mu_\Delta$ & 100.00\% & 0.00\% & $\textcolor{green!60!black}{\downarrow}\,100.00$ & 32.28 & 24.80 & $\textcolor{green!60!black}{\downarrow}\,7.48$ & -- & 1.04 \\
\bottomrule
\end{tabularx}
\end{table}

Circuit searches identify relevant components, but high-fidelity replay requires distributed computation (Appendix~\ref{appendix:circuit}). The vector gives us a compact causal target for tracing how those components form and read out the call decision at the first generated token.

\section{Generalization Beyond Verb-Cued Coding}
\label{sec:generalization}
\label{sec:transfer}

The Qwen3-8B discovery prompts combine a coding task, an explicit request verb, and a single-turn scaffold. We test how the tool-call vector behaves as these conditions change. The tests cover new domains and wording, native multi-turn trajectories, and implicit requests. Every Qwen3-8B intervention keeps the coding-derived direction fixed at the L24 prediction position; cross-domain experiments calibrate its norm, while $\tau^2$ and verb-free tests apply fixed gains to the original vector.

\subsection{Transfer Across Tools, Domains, and Action Wording}
\label{sec:generalization:domains}

We test transfer to web retrieval, read-only SQL execution, and email dispatch using 100 held-out call/no-call pairs per domain. The prompts retain the discovery scaffold and introduce new task content and tool schemas. Call-side verbs such as \emph{retrieve}, \emph{execute}, and \emph{dispatch} fall outside the coding action vocabulary. Within each pair, only the leading request token changes. We keep the coding-derived direction fixed at the L24 prediction position and rescale the vector to match the call--no-call contrast norm estimated from separate target-domain training pairs.

Vector addition induces tool calls on all corrupt prompts, and vector removal suppresses tool calls on clean prompts (\autoref{tab:crossdomain-transfer}). The normalized logit shift averages 0.88 for addition and 0.94 for removal. A norm-matched random vector changes no top-1 decisions. The coding-derived vector supports bidirectional control of the first-token call decision across tools and action vocabularies.

\begin{table}[hb]
\centering
\small
\setlength{\tabcolsep}{3.0pt}
\renewcommand{\arraystretch}{1.05}
\caption{\textbf{The coding-derived vector controls tool calling across web retrieval, SQL execution, and email dispatch.} Qwen3-8B interventions keep the coding-derived direction fixed at L24 and match the vector norm to each target domain. $N$ counts held-out pairs. Induction and suppression rates measure first-token switches from no-call to call under addition and from call to no-call under removal, respectively. Norm.\ shift is the mean \texttt{<tool\_call>} logit increase for Add or decrease for Remove, divided by the baseline call--no-call logit gap. Means weight domains equally.}
\label{tab:crossdomain-transfer}

\begin{tabularx}{\linewidth}{
@{}
>{\hsize=0.95\hsize\raggedright\arraybackslash}X
>{\hsize=1.05\hsize\raggedright\arraybackslash}X
>{\centering\arraybackslash}p{0.65cm}
>{\centering\arraybackslash}p{2cm}
>{\centering\arraybackslash}p{2cm}
>{\centering\arraybackslash}p{2cm}
>{\centering\arraybackslash}p{2cm}
@{}
}
\toprule
& & & \multicolumn{2}{c}{Add $\mu_\Delta$} & \multicolumn{2}{c}{Remove $\mu_\Delta$} \\
\cmidrule(lr){4-5}\cmidrule(lr){6-7}
Target & Tool & $N$ & \shortstack{Induction (\%)} & \shortstack{Norm. shift} & \shortstack{Suppression (\%)} & \shortstack{Norm. shift} \\
\midrule
Web retrieval & \texttt{web\_search} & 100 & 100.0 & 0.93 & 100.0 & 0.75 \\
SQL execution & \texttt{run\_sql} & 100 & 100.0 & 0.79 & 100.0 & 0.77 \\
Email dispatch & \texttt{send\_email} & 100 & 100.0 & 0.91 & 100.0 & 1.30 \\
\midrule
\textbf{Mean} & --- & 300 & \textbf{100.0} & \textbf{0.88} & \textbf{100.0} & \textbf{0.94} \\
\bottomrule
\end{tabularx}
\vspace{-8pt}
\end{table}

\subsection{Native Multi-Turn, Multi-Tool Trajectories}
\label{sec:generalization:tau2}

We next apply the coding-derived vector, without re-estimation, inside ongoing agent interactions. We evaluate native $\tau^2$-Bench Telecom trajectories~\citep{barres2025tau2bench} with 5,385--15,690 context tokens, 16--43 available tools, and prior interaction and call history. We intervene at the next decision point without changing the request, using 200 decision points in each of the removal and induction arms.

At $1\times$ gain, vector removal suppresses 36.7\% of baseline tool calls, and vector addition induces tool calls on 70.0\% of turns with a baseline text response (\autoref{tab:transfer}). A norm-matched random vector switches the call decision on 1.5\% and 0.0\% of these turns, respectively. Across the 200 evaluated decision points in each arm, removal decreases the mean \texttt{<tool\_call>} logit by 5.55, while addition increases it by 19.42. These paired logit changes capture a shift in call preference beyond the decisions whose top-1 token flips. The coding-derived vector therefore remains effective under long contexts, multiple tools, and the accumulated history of interactions and tool calls.

\subsection{Verb-Free Requests}
\label{sec:generalization:implicit}

Requests can also imply a need for tool use without an explicit instruction to act. We construct 600 non-imperative requests across code, search, database, and API tasks. The requests include questions such as \emph{Is it true that \ldots?} and statements such as \emph{I haven't gotten to \ldots\ yet}. We call these requests \emph{verb-free} because they contain no explicit action verb specifying what the model should do.

Among 160 evaluated requests with a baseline Qwen3-8B tool call, vector removal suppresses 93.1\% of calls at $1\times$ gain. At $1.5\times$ gain, the suppression rate reaches 100.0\%, compared with 16.2\% for a norm-matched random vector at the same gain. The paired logit changes in \autoref{tab:transfer} show a decrease in tool-call preference. This supports reuse of the call-state direction on requests that omit the explicit verb cues used to discover $\mu_\Delta$. Appendix~\ref{appendix:generalization:implicit} gives the request construction and selection procedure.

\begin{table}[t]
\centering
\small
\setlength{\tabcolsep}{4pt}
\caption{\textbf{The coding-derived Qwen3-8B vector transfers beyond the discovery prompts.} All entries use $1\times$ gain at L24. $N$ counts evaluated decision points. Top-1 flip is the fraction of baseline calls suppressed by removal or baseline text responses changed to calls by addition. $\Delta z_{\mathrm{call}}$ is the mean paired after-minus-before change in \texttt{<tool\_call>} logit over all $N$ points. Top-1 rate gives the call rate before and after intervention over the same $N$ points. Appendix~\ref{appendix:generalization} gives random controls and shows how the verb-free results vary with the chosen intervention gain.}
\label{tab:transfer}
\label{tab:tau2-qwen8b}
\label{tab:implicit-qwen8b}
\begin{tabularx}{\linewidth}{@{}
>{\raggedright\arraybackslash}X
>{\raggedright\arraybackslash}p{1.35cm}
>{\centering\arraybackslash}p{1.00cm}
>{\centering\arraybackslash}p{2.00cm}
>{\centering\arraybackslash}p{1.50cm}
>{\centering\arraybackslash}p{3.20cm}
@{}}
\toprule
Setting & Intervention & $N$ & \shortstack{Top-1 flip (\%)} & \shortstack{Mean $\Delta z_{\mathrm{call}}$} & \shortstack{Top-1 rate (before $\to$ after)} \\
\midrule
$\tau^2$-Bench, native call & Remove & 200 & 36.7 & $-5.55$  & 99.5\% $\to$ 63.0\% \\
$\tau^2$-Bench, native text & Add   & 200 & 70.0 & $+19.42$ & 0.0\% $\to$ 70.0\%  \\
Verb-free, native call     & Remove & 160 & 93.1 & $-9.95$  & 100.0\% $\to$ 6.9\% \\
\bottomrule
\end{tabularx}
\vspace{-4pt}
\end{table}

\FloatBarrier

\begin{keybox}{Finding: A General Tool-Call Vector Within the Model}
Using controlled verb contrasts and activation patching, we identify a tool-call vector $\mu_\Delta$ within the model. Intervening along this direction controls the call-or-no-call decision across tools and domains, including native multi-turn trajectories and implicit requests.
\end{keybox}
\vspace{-8pt}

\section{How Does the Tool-Call Vector Form?}
\label{sec:formation}

The vector's transfer raises a mechanistic question. How do the scaffold and request create the state captured by $\mu_\Delta$? We return to the controlled Qwen3-8B setting, establish the scaffold's behavioral effect, and trace the vector's formation to individual layers and features.

\subsection{The Scaffold Establishes a Tool-Call Prior}
\label{sec:scaffold-ablation}
\label{sec:scaffold}

The scaffold creates a strong tendency to call tools on task-bearing requests. On held-out tasks, neutral requests produce a mean call probability of 0.8486, while analysis requests reduce the probability to 0.0038. We separate the scaffold into role instructions ($R$), tool schemas ($T$), and format templates ($F$) to identify the source of this tendency. The format template establishes the call prior, while the tool schema makes the resulting call tendency sensitive to the request (\autoref{tab:scaffold-ablation}). Removing $F$ nearly eliminates calling. With $F$ alone, call probabilities approach one for both request types. Removing $T$ weakens the neutral--analysis separation. A length-matched replacement for $T$ produces an intermediate separation, indicating that both schema content and context length contribute.

The prior is gated by the presence of task-relevant content. Empty turns and unrelated questions elicit no top-1 calls. The task body alone yields a 32.3\% call rate. Execution requests reach 100\%, while analysis requests reduce the rate to 0\% in this baseline experiment. We therefore use \emph{scaffold-induced call prior} to denote this task-conditioned tendency to call tools.

\begin{table}[h]
\centering
\small
\caption{\textbf{The scaffold creates a request-sensitive call prior.} Entries are mean first-token call probabilities on 200 held-out Qwen3-8B tasks. Appendix~\ref{appendix:scaffold-controls} gives all component combinations and a control that replaces the tool schema with text of the same length.}
\label{tab:scaffold-ablation}
\begin{tabular}{lrr}
\toprule
Scaffold & Neutral request & Analysis request \\
\midrule
$R+T+F$ (full) & 0.8486 & 0.0038 \\
$R+T$ (no format template) & $6.32\times10^{-7}$ & $8.31\times10^{-10}$ \\
$R+F$ (no tool schema) & 0.9521 & 0.4108 \\
$F$ only & 1.0000 & 0.9952 \\
\bottomrule
\end{tabular}
\end{table}

\subsection{MLPs Dominate Tool-Call Vector Formation}
\label{sec:which}

We track the call signal by projecting residual activations onto the unit direction $\hat{\mu}_\Delta=\mu_\Delta/\|\mu_\Delta\|$. At the prediction position, the mean execution--analysis gap at layer $l$ is
\begin{equation}
  g_l =
  \frac{1}{N}
  \sum_{i=1}^{N}
  \bigl\langle
  h^{(l)}_p(x_i^c)
  -
  h^{(l)}_p(x_i^*),
  \hat{\mu}_\Delta
  \bigr\rangle .
  \label{eq:proj-traj}
\end{equation}
We also project each MLP and attention-head output onto $\hat{\mu}_\Delta$ to measure component contributions.

\begin{figure}[!htbp]
  \centering
  \includegraphics[width=\linewidth]{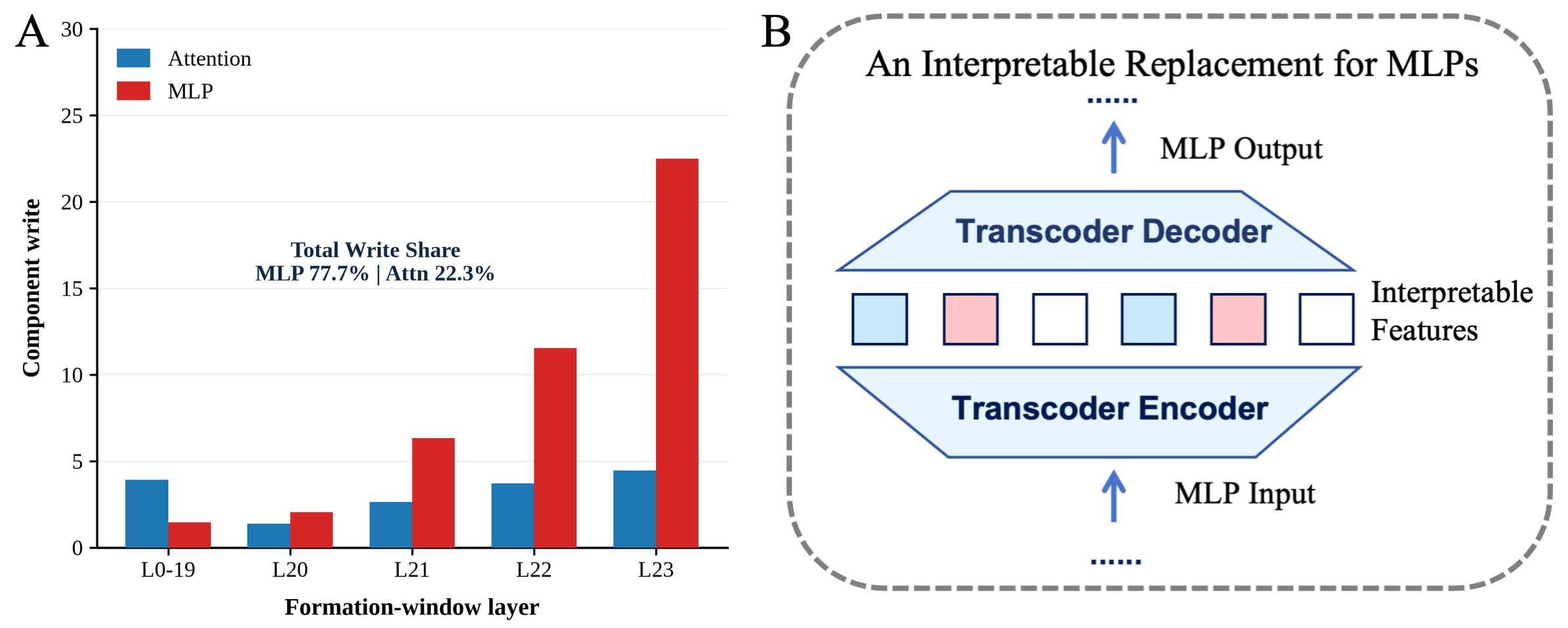}
  \caption{\textbf{MLPs dominate formation-window writes.} Left, Qwen3-8B component outputs projected onto $\hat{\mu}_\Delta$. Right, the Transcoder architecture used to decompose MLP writes.}
  \label{fig:formation}
\end{figure}

The gap $g_l$ stays near zero through the first fifteen layers, grows modestly in the middle layers, and rises sharply in L20--L23. Thus, we call L20--L23 the \emph{formation window}. MLPs supply 77.7\% of the projected write in this window, compared with 22.3\% for attention heads (\autoref{fig:formation}). MLP23 contributes the largest single write, equal to 22.5 on the $g_l$ scale.

\subsection{Suppression Shapes the Vector in the Formation Window}
\label{sec:what-mlps}

We decompose formation-window MLP writes with Transcoders, which approximate each output with feature-weighted decoder directions~\citep{dunefsky2024transcoders}. Feature selection uses 300 training pairs; scores and interventions use 200 held-out pairs and the fixed L24 vector.
\begin{equation}
  \text{MLP}^{(\ell)}(x) \approx b_{\mathrm{dec}}^{(\ell)} + \textstyle\sum_f a_{\ell f}(x)\, w_f^{(\ell)},
  \label{eq:transcoder}
\end{equation}
where $a_{\ell f}(x)\geq 0$ is the ReLU feature activation of the normalized MLP input at the prediction position, and $w_f^{(\ell)}$ is its decoder direction. The bias cancels in paired contrasts. We score each feature by its activation contrast and projection onto $\hat{\mu}_\Delta$,
\begin{equation}
  \kappa_{\ell f} = (\bar{a}^c_{\ell f} - \bar{a}^*_{\ell f})\langle w_f^{(\ell)}, \hat{\mu}_\Delta \rangle,
  \label{eq:kappa}
\end{equation}
where $\bar a^c_{\ell f}$ and $\bar a^*_{\ell f}$ are mean activations on clean and corrupt prompts.

Positive $\kappa$ increases the execution--analysis gap; negative $\kappa$ reduces it. The gap can arise from execution-active features writing along $\hat{\mu}_\Delta$ or analysis-active features writing against it. We sum contribution magnitudes by activation side,
\begin{equation}
    K_{\mathrm{corrupt}}^{(\ell)}
    = \sum_{f:\bar{a}^*_{\ell f}>\bar{a}^c_{\ell f}} |\kappa_{\ell f}|,
    \qquad
    K_{\mathrm{clean}}^{(\ell)}
    = \sum_{f:\bar{a}^c_{\ell f}>\bar{a}^*_{\ell f}} |\kappa_{\ell f}|.
\end{equation}
Let $\beta_{\ell f}=\langle w_f^{(\ell)},\hat{\mu}_\Delta\rangle$. The net contributions of suppressive and promoting writes are $S^{(\ell)}=\sum_{f:\beta_{\ell f}<0}\kappa_{\ell f}$ and $E^{(\ell)}=\sum_{f:\beta_{\ell f}>0}\kappa_{\ell f}$, respectively. Retaining both signs of $\kappa$ gives
\begin{equation}
S^{(\ell)}-E^{(\ell)}
=K_{\mathrm{corrupt}}^{(\ell)}-K_{\mathrm{clean}}^{(\ell)}.
\end{equation}
This identity also holds after summing across layers, so $K_{\mathrm{corrupt}}/K_{\mathrm{clean}}>1$ indicates a larger net suppressive contribution to the execution--analysis gap.

We label each layer's 80 largest positive and 80 largest negative training-$\kappa$ features using max-activating prompts, activation patterns, and decoder tokens. Both the statistics for feature families and those for the full feature set use the same held-out $\kappa$ values.

\begin{table}[h]
\vspace{-3pt}
\centering
\small
\caption{\textbf{Features more active on analysis prompts dominate the formation window.} $K_{\mathrm{corrupt}}$ and $K_{\mathrm{clean}}$ sum $|\kappa|$ over all features with higher analysis and execution activations, respectively, on 200 held-out pairs. Share is each layer's fraction of the net Transcoder feature write over L20--L23. Appendix~\ref{appendix:transcoder} gives the statistics for the labelled families of Transcoder features.}
\label{tab:transcoder-features}
\begin{tabularx}{\linewidth}{clccccX}
\toprule
Layer & Dominant & $K_{\mathrm{corrupt}}$ & $K_{\mathrm{clean}}$ & $K_{\mathrm{corrupt}}/K_{\mathrm{clean}}$ & Share (\%) & Semantic label \\
\midrule
L20 & Clean   & 0.24  & 0.93  & 0.26 & 4.6  & Execution requests \\[2pt]
L21 & Corrupt & 1.47  & 1.10  & 1.34 & 16.7 & Non-necessity \\[2pt]
L22 & Corrupt & 1.85  & 1.33  & 1.39 & 22.8 & Analysis-task contexts \\[2pt]
L23 & Corrupt & 5.98  & 1.43  & 4.18 & 55.9 & Analysis-verbs \\
\bottomrule
\end{tabularx}
\vspace{-3pt}
\end{table}

The scaffold already supports frequent calls on neutral requests (\autoref{sec:scaffold}), providing a behavioral baseline for interpreting the execution--analysis contrast. Across the formation window, analysis-active features carry nearly twice the contribution magnitude of execution-active features (\autoref{tab:transcoder-features}). In the \emph{analysis-non-execution} family, which has the largest total contribution among the labelled families, most features are more active on analysis prompts and write against $\hat{\mu}_\Delta$. Their stronger analysis-side activation therefore enlarges the paired gap through stronger negative writes on analysis prompts. These observations support an asymmetric formation process: execution requests largely preserve the scaffold-induced call tendency, while analysis requests actively suppress it. The execution-minus-analysis vector thus largely reflects the suppression present on the analysis side.

We next examine whether this analysis-side dominance translates into substantial influence over the call decision. If the identified features actively restrain the call tendency, removing their writes should increase calling while the analysis request remains unchanged. Zeroing the writes of the five features with the largest suppressive contributions ($|\kappa|$), comprising less than 0.001\% of the 655,360 Transcoder features across L20--L23, shifts the L24 state along $+\hat{\mu}_\Delta$ and raises the call rate on analysis prompts to 25.5\% (Appendix~\ref{appendix:transcoder}). The increase in calling under unchanged request wording supports a causal role for these features in suppressing the scaffold-induced call tendency.

\Needspace{6\baselineskip}
\begin{keybox}{Mechanistic Insight: Suppression Shapes the Tool-Call Vector}
The scaffold establishes a conditional tool-call prior. Analysis requests activate suppressor features that write against the tool-call direction $\hat{\mu}_\Delta$, while execution requests largely preserve the prior. This asymmetry between request types shapes the tool-call vector.
\end{keybox}

\FloatBarrier

\section{How Does the Tool-Call Vector Control the Call Decision?}
\label{sec:readout}

Downstream attention heads and MLP features read the L24 tool-call state into the first generated token. We trace this readout with direct logit attribution (DLA) and single-component patching over L25--L35. DLA measures each component's direct contribution to the tool-call logit~\citep{elhage2021framework}.

\subsection{Scaffold-Reading Heads Support Tool Calling}
\label{sec:attn-readout}

\begin{figure}[!htbp]
  \vspace{-6pt}
  \centering
  \includegraphics[width=0.9\textwidth]{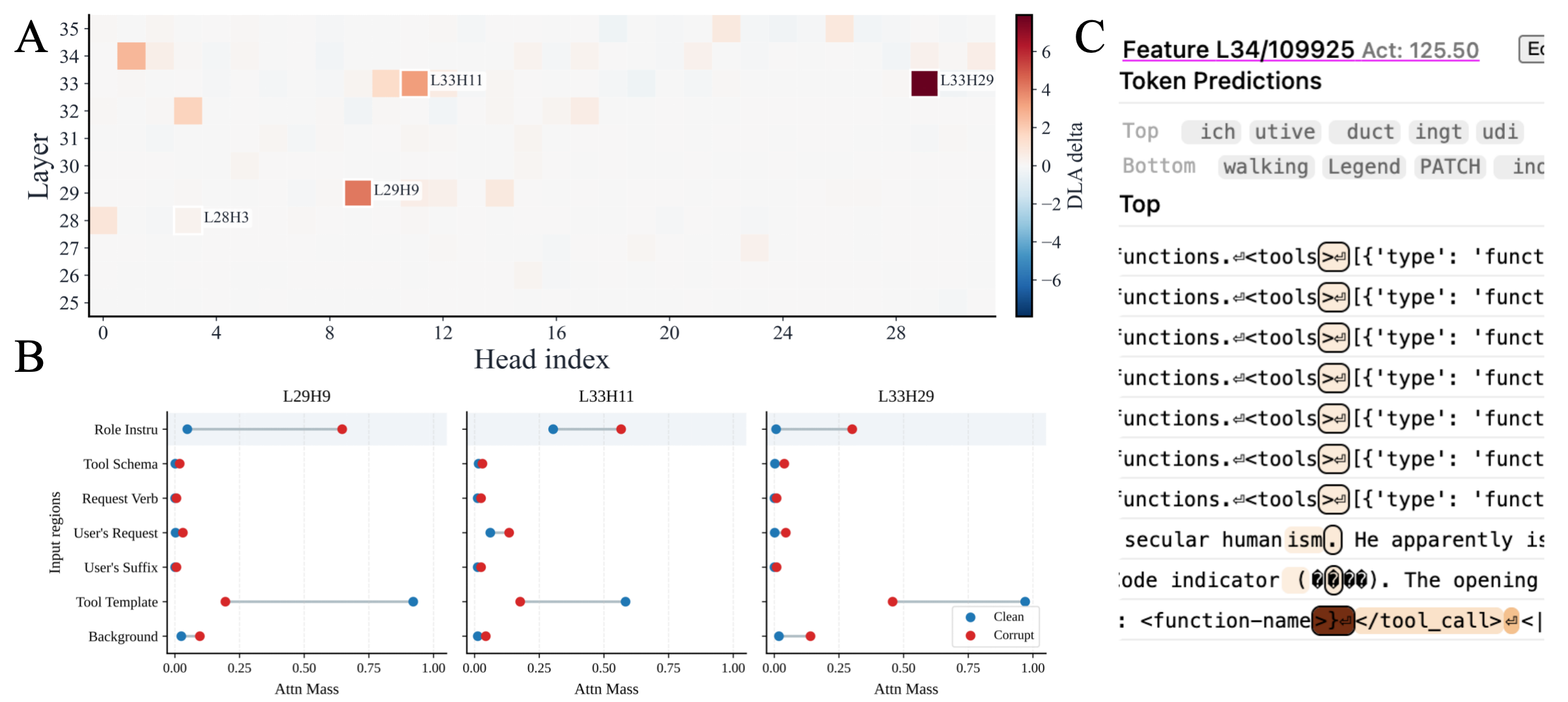}
  \caption{\textbf{Scaffold-reading heads and a late structural feature read out the tool-call state.} (A) Execution-minus-analysis DLA for the \texttt{<tool\_call>} logit across Qwen3-8B attention heads in L25--L35. (B) L29H9, L33H11, and L33H29 share a shift toward the tool-format template and away from role instructions on execution prompts (blue) relative to analysis prompts (red). (C) Max-activating contexts for L34/F109925 highlight tool-schema boundaries, supporting its interpretation as a structural readout feature for the opening of a tool call.}
  \label{fig:readout}
  \vspace{-6pt}
\end{figure}

To identify candidate readers of the L24 call state, we compare downstream attention heads' direct logit attribution (DLA) to \texttt{<tool\_call>} on execution and analysis prompts. L29H9, L33H11, and L33H29 combine stronger execution-side call-token support with a shared shift of attention from role instructions toward the tool-format template (\autoref{fig:readout}A--B). We therefore refer to them as \emph{scaffold-reading heads}. We then test whether their call-directed writes respond to the internal state by adding $\mu_\Delta$ at L24 while leaving the analysis prompt unchanged. This intervention brings all three heads' DLA close to their execution-side values (\autoref{tab:appendix-readout-vector-response}). The recovered output support under fixed prompt content shows that the internal call state regulates these heads' call-directed writes, supporting their role in reading available scaffold information into the call-opening token.

\subsection{Committing to the Tool-Call Opening}
\label{sec:mlp-readout}

MLP34's strong patching effect motivates a feature-level test of how the call state is expressed in the output format. Transcoder feature F109925 responds to tool-schema boundary contexts and writes toward the call-opening token (\autoref{fig:readout}C). Adding $\mu_\Delta$ at L24 brings both its activation and its call-token write on analysis prompts close to their execution-side values, linking the feature's activity to the internal call state. To test whether restoring this feature also changes the decision, we replace its analysis-side activation with the paired execution-side value at the prediction position. On the same 200 held-out pairs, this yields 37.0\% calling, compared with 6.5\% for the fixed non-structural control F91365 and 41.0\% for the full MLP34 patch (Appendix~\ref{appendix:readout}). Its response to $\mu_\Delta$ and the behavioral recovery from feature replacement together support F109925 as a causal contributor to expressing the call state in the token that initiates tool use.

\begin{keybox}{Takeaway: From Scaffold Prior to Tool-Call Output}
The scaffold establishes a conditional tool-call prior. Analysis requests activate suppressor features that write against the tool-call direction $\hat{\mu}_\Delta$, while execution requests preserve the prior. Scaffold-reading heads and late structural MLP features then read the resulting state into the \texttt{<tool\_call>} opening as the first token of the response.
\end{keybox}

\section{The Tool-Call Vector Across Scales and Model Families}
\label{sec:crossscale}

We ask whether the mechanism established in Qwen3-8B replicates across model scales and families, and whether the resulting vectors generalize beyond the discovery prompts: (i) the decision state localizes to a prediction-position layer ($l$, $r(l,p)$); (ii) $\mu_\Delta$ passes causal intervention tests at that layer (Suff., Necc.); (iii) the vector generalizes beyond the controlled verb contrasts to new domains, native multi-turn $\tau^2$-Bench trajectories, and verb-free requests (Domains, $\tau^2$, Verb-free); (iv) formation-window writes are MLP-dominated and suppressor-driven (MLP/Attn, $K_{\mathrm{corrupt}}/K_{\mathrm{clean}}$); and (v) scaffold attention varies with the call state (Max Attn). We test Qwen3-4B and 14B, Qwen3.5-4B and 9B, Mistral-Small-3.2-24B, and Granite-3.3-8B at the fixed block-input layers in \autoref{tab:crossscale}.

\begin{table}[h]
\centering
\scriptsize
\caption{\textbf{The tool-call mechanism recurs across seven models and generalizes beyond the discovery prompts.} Domains and $\tau^2$ average conditional intervention rates; Verb-free reports suppression at unit gain. Domains uses norm calibration (Appendix~\ref{appendix:implementation:compute}). Max Attn summarizes clean-minus-corrupt scaffold-attention shifts (Appendix~\ref{appendix:cross-scale:readers}). For hybrid Qwen3.5, MLP/Attn is omitted and attention is summarized only for layers with full attention.}
\label{tab:crossscale}

\setlength{\tabcolsep}{2.3pt}
\resizebox{\textwidth}{!}{%
\begin{tabular}{l cc cc ccc cc c}
\toprule
& \multicolumn{2}{c}{Localization}
& \multicolumn{2}{c}{Intervention}
& \multicolumn{3}{c}{Generalization (\%)}
& \multicolumn{2}{c}{Formation}
& Readout \\
\cmidrule(lr){2-3}
\cmidrule(lr){4-5}
\cmidrule(lr){6-8}
\cmidrule(lr){9-10}
\cmidrule(lr){11-11}
Model & $l$ & $r(l,p)$ (\%) & Suff. & Necc. & Domains & $\tau^2$ & Verb-free & MLP/Attn & $K_{\mathrm{corrupt}}/K_{\mathrm{clean}}$ & Max Attn (pp) \\
\midrule
Qwen3-4B & 26 & 100.0 & 0.97 & 0.71 & 87.7 & 72.2 & 100.0 & 2.08 & 3.26 & 82.6 \\
Qwen3-8B        & 24 & 100.0 & 1.03 & 1.04 & 100.0 & 53.3 & 93.1  & 3.48 & 1.99 & 72.6 \\
Qwen3-14B & 34 & 100.0 & 0.97 & 0.84 & 100.0 & 85.0 & 100.0 & 6.87 & 2.41 & 87.3 \\
Qwen3.5-4B & 31 & 99.0 & 0.84 & 0.61 & 91.1 & 47.9 & 100.0 & -- & 1.58 & 50.3 \\
Qwen3.5-9B & 31 & 100.0 & 0.92 & 0.88 & 100.0 & 80.2 & 100.0 & -- & 1.30 & 52.2 \\
Mistral-Small-3.2-24B & 25 & 100.0 & 0.90 & 0.76 & 100.0 & 83.5 & 100.0 & 3.93 & 1.61 & 40.5 \\
Granite-3.3-8B & 35 & 93.5 & 0.81 & 0.72 & 43.0 & 70.3 & 35.5 & 7.52 & 2.11 & 36.5 \\
\bottomrule
\end{tabular}%
}
\end{table}

Across all seven models, localization and causal intervention remain consistently strong, while the strength of generalization and downstream readout varies across architectures. Full-state patching recovers calls on 93.5--100.0\% of analysis prompts. Vector interventions yield normalized sufficiency scores of 0.81--1.03 and necessity scores of 0.61--1.04. More importantly, $K_{\mathrm{corrupt}}>K_{\mathrm{clean}}$ in every model, preserving the central asymmetry identified in Qwen3-8B: the tool-call vector is shaped primarily by suppressive computation on the analysis side rather than by an execution-specific signal. Together, these results suggest that the specific layers and components vary across models, while the functional organization of the call decision remains stable across these models.

\section{Related Work}
\label{sec:relatedwork}

Tool-use research develops and evaluates models that decide whether and how to invoke external tools~\citep{schick2023toolformer,patil2023gorilla,yao2023react,huang2024metatool,patil2025bfcl}. These studies establish the call-or-no-call decision as a central agent capability. Its causal internal representation remains largely uncharacterized. Mechanistic interpretability localizes causal states and resolves component computations~\citep{meng2022locating,heimersheim2024patching,conmy2023acdc,dunefsky2024transcoders}, while representation engineering, function vectors, and refusal directions show that compact residual directions can mediate behavior~\citep{zou2023repe,todd2024functionvectors,arditi2024refusal}. We connect these lines of work by extending causal mechanistic analysis to a core agent behavior: deciding whether to initiate tool use. Controlled request contrasts isolate this decision in long, scaffolded prompts, revealing a localized tool-call vector that transfers across tools, domains, and request forms. Tracing its formation and downstream readout explains how request semantics suppress a scaffold-induced call prior, linking behavioral tool-use evaluation to an internal causal account of action initiation. Appendix~\ref{appendix:relatedwork} provides a review of related literature.

\section{Conclusions}
\label{sec:discussion}

A single tool-call vector gives causal control over the call-or-no-call decision at the first output token in Qwen3-8B. The vector is estimated from one-word request contrasts and transfers without re-estimation to native multi-turn trajectories and verb-free requests. The mechanistic analysis explains how the controlled contrast arises. The scaffold establishes a call prior on task-bearing requests, analysis requests activate suppressor features, and execution requests largely preserve the prior. Downstream attention and MLP features read the resulting state into the tool-call opening. The vector connects these computations through a compact, reusable representation, providing a concrete account of how the model uses request semantics to regulate tool calling.

\section{Limitations and Broader Impacts}
\label{sec:limitations-impacts}

Our scope is the mechanism of the binary call-or-no-call decision. The primary formation analysis uses behaviorally screened coding pairs under a fixed scaffold, so its applicability to other prompt constructions remains to be established. We do not evaluate the broader correctness of tool use, including whether a call selects an appropriate tool, produces valid arguments, or completes the underlying task. The formation analysis identifies a scaffold-induced call prior, a vector, and broad suppressive feature families. It does not yet resolve the fine-grained computation that maps scaffold information and request semantics to the vector. Transfer supports reuse of a downstream call state, but does not establish identical formation mechanisms for explicit and implicit requests. Our findings may help developers understand and regulate when AI agents use tools. The same interventions could also induce tool calls that conflict with a user's intent.

\newpage
\bibliographystyle{unsrtnat}
\bibliography{references}

\newpage
\newpage

\appendix
\renewcommand{\ttdefault}{lmtt}

\section{Experimental Setup and Dataset Construction}
\label{appendix:dataset}
This appendix describes the model-specific paired-prompt datasets and the
controls that support the Tool-Call Vector. Each model in the seven-model
comparison has 500 clean/corrupt pairs. Within a pair, the task or conversation and tool interface stay fixed
while the request verb changes. Call outcomes are measured at the first output
token.

\subsection{Task Sources for Each Model's Paired Prompts}
\label{appendix:dataset:sources}

The model-specific corpora use programming tasks from APPS, CodeContests,
HumanEval, and MBPP, or conversation contexts from $\tau^2$-Bench
Telecom. Source families vary by model; Table~\ref{tab:appendix-source-tasks}
summarizes the assignment.

\begin{table}[h]
\centering
\small
\caption{Source families represented in the model-specific paired corpora.}
\label{tab:appendix-source-tasks}
\begin{tabularx}{\linewidth}{>{\raggedright\arraybackslash}X>{\raggedright\arraybackslash}X>{\raggedright\arraybackslash}X}
\toprule
Source family & Models & Task content \\
\midrule
APPS, CodeContests, HumanEval, MBPP & Qwen3-4B, Qwen3-8B, Qwen3-14B, Qwen3.5-9B, Granite-3.3-8B & Programming tasks with statements, function signatures, examples, tests, or reference solutions. \\
$\tau^2$-Bench Telecom & Qwen3.5-4B, Mistral-Small-3.2-24B & Conversation histories with a controlled request-verb contrast. \\
\bottomrule
\end{tabularx}
\end{table}

For coding tasks, normalization preserves the function behavior, programming
language, and available examples or tests. The paired prompts keep each
model's task context and tool interface fixed while varying the request verb.

\subsection{Qwen3-8B Prompt Template}
\label{appendix:dataset:template}

The Qwen3-8B discovery prompts use Qwen's chat template. The system turn
provides the \texttt{write\_file} tool and states the required \texttt{<tool\_call>}
serialization format. The assistant prefix uses the non-thinking completion
form used by Qwen3. The boxed template below shows the structure; braces mark
fields filled by each task. Cross-model corpora use each model's native prompt
template and call-opening token.

\Needspace{22\baselineskip}
\promptboxcaption{Agentic prompt template for the controlled Qwen3 coding pairs.}
\begin{PromptBox}
<|im_start|>system
# Tools

You may call one or more functions to assist with the user query.

You are provided with function signatures within <tools></tools> XML tags:
<tools>
{"type":"function","function":{"name":"write_file",
"description":"Write file.",
"parameters":{"type":"object","properties":{"file_path":{"type":"string"},
"content":{"type":"string"}},"required":["file_path","content"]}}}
</tools>

For each function call, return a json object with function name
and arguments within <tool_call></tool_call> XML tags:
<tool_call>
{"name": <function-name>, "arguments": <args-json-object>}
</tool_call><|im_end|>
<|im_start|>user
{REQUEST_VERB} the function body in {TARGET_FILE} based on the function
definition and docstring below:
{TASK_BODY}
<|im_end|>
<|im_start|>assistant
<think>

</think>
\end{PromptBox}

The tool list, tool schema, output format instruction, target file, task body,
and assistant prefix are identical within each pair; the request verb at the
beginning of the user turn is the only change.

The call marker is \texttt{<tool\_call>} for Qwen3 and Qwen3.5,
\texttt{[TOOL\_CALLS]} for Mistral, and \texttt{<|tool\_call|>} for Granite.
The evaluation compares the native token's logit with all competing first
tokens. For inputs stored as native token IDs, evaluation uses those IDs directly.

\subsection{Pair Selection and Data Splits}
\label{appendix:dataset:pipeline}

The paired corpora are organized separately by model. Pairs are retained
for an execution-call/analysis-text behavioral contrast. Each model has 500
pairs, with the split and source family shown in
Table~\ref{tab:appendix-dataset-splits}. Verb pools are model-specific.
Vector fitting uses training pairs only, and intervention results are
evaluated on held-out pairs. The main paired-prompt experiments use 300
training pairs and 200 held-out pairs for every model. Additional appendix
controls use the evaluation cohorts specified in their respective subsections.

\begin{table}[h]
\centering
\small
\caption{Per-model paired-prompt corpus sizes and splits.}
\label{tab:appendix-dataset-splits}
\begin{tabular}{llrr}
\toprule
Model & Source family & Train & Held-out \\
\midrule
Qwen3-4B & Coding tasks & 300 & 200 \\
Qwen3-8B & Coding tasks & 300 & 200 \\
Qwen3-14B & Coding tasks & 300 & 200 \\
Qwen3.5-4B & $\tau^2$-Bench Telecom & 300 & 200 \\
Qwen3.5-9B & Coding tasks & 300 & 200 \\
Mistral-Small-3.2-24B & $\tau^2$-Bench Telecom & 300 & 200 \\
Granite-3.3-8B & Coding tasks & 300 & 200 \\
\bottomrule
\end{tabular}
\end{table}

\subsection{Qwen3-8B Dataset Composition}
\label{appendix:dataset:results}

The Qwen3-8B corpus contains 500 pairs, split into 300 training and 200
held-out pairs. Its verb counts by split are given in
Table~\ref{tab:appendix-verb-distribution}; the source composition is given
in Table~\ref{tab:appendix-qwen8-source-mix}.

\begin{table}[h]
\centering
\small
\caption{Qwen3-8B execution- and analysis-verb counts by split.}
\label{tab:appendix-verb-distribution}
\begin{tabularx}{\linewidth}{l>{\raggedright\arraybackslash}X>{\raggedright\arraybackslash}X}
\toprule
Split & Execution verbs & Analysis verbs \\
\midrule
Train (300) & add 60, build 60, complete 60, save 60, write 60 & discuss 71, explore 75, inspect 6, review 72, study 76 \\
Held-out (200) & add 53, build 20, complete 20, save 53, write 54 & discuss 40, explore 40, inspect 40, review 40, study 40 \\
\bottomrule
\end{tabularx}
\end{table}

\begin{table}[h]
\centering
\small
\caption{Source composition of the Qwen3-8B paired-prompt corpus.}
\label{tab:appendix-qwen8-source-mix}
\begin{tabular}{lrr}
\toprule
Source & Train & Held-out \\
\midrule
APPS & 192 & 51 \\
CodeContests & 63 & 135 \\
HumanEval & 15 & 4 \\
MBPP & 30 & 10 \\
\midrule
Total & 300 & 200 \\
\bottomrule
\end{tabular}
\end{table}

\Needspace{20\baselineskip}
\subsection{A Representative Clean/Corrupt Prompt Pair}
\label{appendix:dataset:example}

The example below instantiates the task-bearing user turn and assistant prefix of the template in Appendix~\ref{appendix:dataset:template}. The paired analysis request changes \emph{Save} to \emph{Explore}; the shared system turn and tool schema follow that template.

\promptboxcaption{Representative execution request and completion prefix, following the shared system turn.}
\begin{PromptBox}
<|im_start|>user
Save the function body in solve.py based on the function definition and
docstring below:
def isLongPressedName(name: str, typed: str) -> bool:
    """
    Given strings name and typed, return whether typed could result from
    typing name where characters may be long-pressed and repeated one or
    more times.
    >>> isLongPressedName('alex', 'aaleex')
    True
    """
    pass
<|im_end|>
<|im_start|>assistant
<think>

</think>
\end{PromptBox}

\promptboxcaption{Paired analysis request. All remaining prompt content is unchanged.}
\begin{PromptBox}
Explore the function body in solve.py based on the function definition and
docstring below:
\end{PromptBox}

\section{Identifying the Tool-Call Vector}
\label{appendix:vector}
The main text localizes the call-or-no-call state to the prediction position and
summarizes it with the mean clean--corrupt residual difference
$\mu_\Delta$. The controls below test whether the causal effect depends on the
selected layer and direction. They use the same first-token call metric as the
main experiments and separate the learned orientation from perturbation norm.

\subsection{L24 Supports Both Call Induction and Suppression}
\label{appendix:early-layer:controls}

The neighboring-layer control tests how the causal effect depends on the
intervention layer. We fit a mean-difference direction at the input residual
stream of L22, L23, L24, and L25 using 200 training pairs and evaluate each
direction on 300 disjoint held-out pairs from a separate control corpus.
This neighboring-layer control uses a 200/300 split; the main Qwen3-8B
vector uses the 300/200 split above. Each learned vector is compared with
an equal-norm random direction at the same layer. Matching layer and norm
isolates the contribution of the learned orientation.

\begin{table}[h]
\centering
\small
\caption{Neighboring-layer control with independently fitted directions.
Entries are held-out post-intervention call rates. Random controls match
the learned direction's norm at each layer.}
\label{tab:appendix-layer-controls}
\begin{tabular}{lrrrr}
\toprule
Intervention & L22 & L23 & L24 & L25 \\
\midrule
Analysis $+\mu_\Delta$ & 82.0\% & 100.0\% & 100.0\% & 100.0\% \\
Execution $-\mu_\Delta$ & 45.7\% & 12.0\% & 0.0\% & 0.0\% \\
Analysis $+$ random & 0.0\% & 0.3\% & 18.7\% & 0.0\% \\
Execution $-$ random & 100.0\% & 99.7\% & 97.7\% & 100.0\% \\
\bottomrule
\end{tabular}
\end{table}

The addition and removal arms reveal complementary aspects of call control. Induction becomes complete at L23, while complete suppression appears at L24. L24 is therefore the earliest tested layer that combines both effects, which also persist at L25 (Table~\ref{tab:appendix-layer-controls}). The weaker effects of norm-matched random directions link this control to the fitted orientation. Together with the full-state patching result, this makes L24 a well-supported reference state for tracing formation and downstream readout.

\section{Generalization and Transfer}
\label{appendix:generalization}
The paired prompts hold each model's task context and tool interface fixed while varying the request verb. Transfer tests whether the resulting call-control direction remains effective across task domains, tool schemas, contexts, and request wording. Each model retains its fitted direction throughout; target-domain norm calibration adjusts its magnitude using separate training pairs.

\subsection{Call Induction and Suppression in Native \texorpdfstring{$\tau^2$}{tau2}-Bench Trajectories}
\label{appendix:generalization:tau2}

We evaluate $\tau^2$-Bench~\citep{barres2025tau2bench} trajectories at each model's next native decision point: Telecom for the Qwen models and Retail for Mistral and Granite. Each arm begins with 200 candidate decision points. Removal is evaluated on the $N_{\mathrm{call}}$ points whose baseline first token is the model's native call marker; induction is evaluated on $N_{\mathrm{text}}=200$ baseline text decisions. We apply the fitted $\mu_\Delta$ and a norm-matched random direction at unit gain. Tables~\ref{tab:appendix-tau2-removal} and~\ref{tab:appendix-tau2-induction} report strict first-token flip rates for the two arms; the removal table gives each model's baseline-call denominator.

\begin{table}[h]
\centering
\small
\caption{$\tau^2$-Bench removal: strict first-token flip rate among the $N_{\mathrm{call}}$ native baseline tool-call decisions.}
\label{tab:appendix-tau2-removal}
\begin{tabular}{lrcc}
\toprule
Model & $N_{\mathrm{call}}$ & $\mu_\Delta$, $1\times$ & Random, $1\times$ \\
\midrule
Qwen3-4B & 199 & 61.3\% & 0.5\% \\
Qwen3-8B & 199 & 36.7\% & 1.5\% \\
Qwen3-14B & 200 & 97.0\% & 8.0\% \\
Qwen3.5-4B & 198 & 88.4\% & 0.0\% \\
Qwen3.5-9B & 196 & 91.3\% & 17.3\% \\
Mistral-Small-3.2-24B & 200 & 75.0\% & 0.5\% \\
Granite-3.3-8B & 200 & 95.5\% & 2.0\% \\
\bottomrule
\end{tabular}
\end{table}

\begin{table}[h]
\centering
\small
\caption{$\tau^2$-Bench induction: strict first-token flip rate among native
text decisions.}
\label{tab:appendix-tau2-induction}
\begin{tabular}{lcc}
\toprule
Model & $\mu_\Delta$, $1\times$ & Random, $1\times$ \\
\midrule
Qwen3-4B & 83.0\% & 0.0\% \\
Qwen3-8B         & 70.0\% & 0.0\% \\
Qwen3-14B & 73.0\% & 0.0\% \\
Qwen3.5-4B & 7.5\% & 0.0\% \\
Qwen3.5-9B & 69.0\% & 0.0\% \\
Mistral-Small-3.2-24B & 92.0\% & 0.0\% \\
Granite-3.3-8B & 45.0\% & 0.0\% \\
\bottomrule
\end{tabular}
\end{table}

Transfer in native trajectories tests whether the discovery direction controls call initiation as the surrounding task, history, and tool set change. The fitted directions outperform norm-matched random perturbations in both arms (Tables~\ref{tab:appendix-tau2-removal} and~\ref{tab:appendix-tau2-induction}). Each intervention leaves the native request and conversation fixed, so the resulting flip connects the transferred internal state to the call decision in that interaction. Induction and removal establish this connection from both baseline text and baseline call states, extending the direction's action-initiation role to ongoing conversations with different tools and histories.

\subsection{Call Suppression on Verb-Free Requests Across Models}
\label{appendix:generalization:implicit}

We construct 600 verb-free requests from APPS
\citep{hendrycks2021apps}, FEVER~\citep{thorne2018fever}, Spider
\citep{yu2018spider}, and an API domain, using five non-imperative phrasing
patterns with 30 requests per domain--pattern combination. Requests are
rendered in each model's native template. After evaluating the baseline on
all 600 requests, we select up to ten tool-call-positive requests per
domain--pattern combination in source order. This subset is fixed across
the removal and random-control interventions. We subtract $\mu_\Delta$ at
gains $\alpha \in \{1,1.5\}$
and compare against a norm-matched random direction at $\alpha=1.5$.
\autoref{tab:appendix-implicit-full} reports the strict drop rate of each
model's native tool-call marker among these baseline-positive requests.

\begin{table}[h]
\centering
\small
\caption{Implicit, verb-free generalization across all seven models. $N$ is
the evaluated baseline-positive subset; $r$ is a norm-matched random direction.}
\label{tab:appendix-implicit-full}
\begin{tabular}{lcccc}
\toprule
Model & $N$ & $(-\mu_\Delta)$, $\alpha=1$ & $(-\mu_\Delta)$, $\alpha=1.5$ & $(-r)$, $\alpha=1.5$ \\
\midrule
Qwen3-4B & 117 & 100.0\% & 100.0\% & 34.2\% \\
Qwen3-8B              & 160 & 93.1\%  & 100.0\% & 16.2\% \\
Qwen3-14B & 161 & 100.0\% & 100.0\% & 44.7\% \\
Qwen3.5-4B & 186 & 100.0\% & 100.0\% & 33.9\% \\
Qwen3.5-9B & 187 & 100.0\% & 100.0\% & 73.8\% \\
Mistral-Small-3.2-24B & 147 & 100.0\% & 100.0\% & 5.4\% \\
Granite-3.3-8B & 200 & 35.5\% & 88.5\% & 27.5\% \\
\bottomrule
\end{tabular}
\end{table}

Removing the fitted direction suppresses calls even when requests omit the explicit verb cues used to discover it. The same direction therefore controls call initiation across explicit and implicit request forms (Table~\ref{tab:appendix-implicit-full}). The gain response further shows how this control depends on displacement along the direction: increasing the gain raises Granite's suppression from 35.5\% to 88.5\%. The learned-direction advantage over norm-matched random perturbations at the same gain connects this effect to the fitted orientation. These results support a reusable downstream call state reached through different request formulations.

\section{Formation of the Tool-Call Vector}
\label{appendix:early-layer}
\label{appendix:formation}
The formation analysis asks how the scaffold and request wording produce the state identified in Appendix~\ref{appendix:vector}. Scaffold and affordance controls establish when a call tendency is prepared and how requests regulate it. We then connect request information, intermediate call control, and output support, before testing the feature writes that form the call state.

\subsection{The Scaffold Establishes a Task-Conditioned Call Prior}
\label{appendix:scaffold-controls}

We ablate role instructions ($R$), tool schemas ($T$), and format templates
($F$) on the 200 held-out tasks in the balanced Qwen3-8B dataset. A
length-matched control replaces the tool schema $T$, while retaining
$R$ and $F$. The neutral requests use a balanced assignment of
\emph{Consider}, \emph{Handle}, \emph{Take}, \emph{Use}, and
\emph{Process}. Table~\ref{tab:appendix-scaffold-factorial} gives the
component ablations and the length-matched control.

\begin{table}[h]
\centering
\small
\caption{Scaffold-component ablation, Qwen3-8B, 200 held-out prompts per condition. $p_{\mathrm{call}}$ is the mean first-token \texttt{<tool\_call>} probability; $P$ is the neutral-minus-analysis gap.}
\label{tab:appendix-scaffold-factorial}
\begin{tabularx}{\linewidth}{>{\raggedright\arraybackslash}Xccc}
\toprule
Scaffold components & Neutral $p_{\mathrm{call}}$ & Analysis $p_{\mathrm{call}}$ & $P$ \\
\midrule
$R+T+F$ (full scaffold) & 0.8486 & 0.0038 & 0.845 \\
$R+F$ (no tool schema) & 0.9521 & 0.4108 & 0.541 \\
$R+T$ (no format template) & $6.32\times10^{-7}$ & $8.31\times10^{-10}$ & $6.3\times10^{-7}$ \\
$F$ only & 1.0000 & 0.9952 & 0.005 \\
$T$ only & $1.87\times10^{-8}$ & $2.05\times10^{-11}$ & $1.9\times10^{-8}$ \\
$R$ only & $8.77\times10^{-9}$ & $7.14\times10^{-11}$ & $8.7\times10^{-9}$ \\
Empty system scaffold & $6.80\times10^{-16}$ & $1.05\times10^{-15}$ & $-3.7\times10^{-16}$ \\
$R$ + length-matched neutral text + $F$ & 0.7311 & 0.1553 & 0.576 \\
\bottomrule
\end{tabularx}
\end{table}

The factorial ablations separate a high call tendency from sensitivity to the request. With $F$ alone, both request types strongly favor the call opening; removing $F$ nearly eliminates that tendency. With role instructions and the format template held fixed, including $T$ strengthens the neutral--analysis separation. Replacing $T$ with length-matched neutral text weakens this separation even though context length is retained, supporting a role for schema content beyond token count (Table~\ref{tab:appendix-scaffold-factorial}). The residual separation under replacement shows that the retained scaffold and request wording also contribute to request sensitivity.

The separate 300-task request cohort shows that the scaffold establishes a call tendency conditional on task context (Table~\ref{tab:appendix-request-baselines}). Neutral wording already elicits frequent calls; execution wording reinforces this tendency, while analysis wording suppresses it. Empty, greeting, and unrelated turns keep calling inactive. These controls place the asymmetry in how the request regulates a task-conditioned prior: execution requests preserve or strengthen a prepared action, and analysis requests override it.

\begin{table}[h]
\centering
\small
\caption{Qwen3-8B request baselines on a 300-task evaluation cohort. Empty,
greeting, and unrelated turns are evaluated once each. The full scaffold is
present except in the final row.}
\label{tab:appendix-request-baselines}
\begin{tabular}{lrr}
\toprule
Request & Call top-1 & Mean call probability \\
\midrule
Empty turn, greeting, or unrelated question & 0.0\% & $\approx 5.0\times10^{-4}$ (max.) \\
Task body only & 32.3\% & 0.3310 \\
Neutral verb and task body & 86.7\% & 0.8547 \\
Analysis verb and task body & 0.0\% & 0.0080 \\
Execution verb and task body & 100.0\% & 0.9993 \\
Execution verb and task body, no scaffold & 0.0\% & $6.56\times10^{-17}$ \\
\bottomrule
\end{tabular}
\vspace{-4pt}
\end{table}

\subsection{Tool Affordances Change Whether a Request Elicits a Call}
\label{appendix:affordance}

The available tool changes what it means to act on a request. We cross \emph{Write} and \emph{Review} with \texttt{write\_file} and \texttt{submit\_review}, using 100 requests per condition. The \emph{Review} request elicits 4.0\% calling with the writing tool and 100.0\% when review submission is available (Table~\ref{tab:appendix-affordance}). Review becomes an actionable instruction when the interface provides the corresponding operation. The execution/analysis contrast therefore depends on the relation between request meaning and tool affordances, which gives the discovery verb pools their behavioral significance in that interface.

\begin{table}[h]
\centering
\small
\caption{Request--tool crossover on Qwen3-8B. Entries are tool call
rates, with 100 requests.}
\label{tab:appendix-affordance}
\begin{tabular}{lrr}
\toprule
Request verb & \texttt{write\_file} & \texttt{submit\_review} \\
\midrule
Write & 100.0\% & 34.0\% \\
Review & 4.0\% & 100.0\% \\
\bottomrule
\end{tabular}
\vspace{-4pt}
\end{table}

\subsection{Request Encoding Precedes Formation and Readout of the Call State}
\label{appendix:early-separability}
We ask how the timing of request encoding relates to formation of the call state and its expression in the output token. At the prediction position, a linear probe measures whether the residual stream distinguishes execution from analysis requests, the logit lens measures the call-token preference, and MLP DLA identifies direct writes toward that token. The probe first exceeds 0.9 AUC at L1, the call-token logit-lens gap first exceeds 1.0 at L27, and the largest MLP DLA differences occur at L33--L35 (Table~\ref{tab:early-signal-triplet}). Thus, the request distinction is available early, while its strongest direct expression in call-token writes occurs near the output.

The formation and intervention results explain the role of the computation between these timing landmarks. Formation-window MLPs build the execution--analysis gap along $\hat{\mu}_\Delta$ in L20--L23 (\autoref{sec:which}). At L24, adding $\mu_\Delta$ on analysis prompts induces calling, and subtracting it on execution prompts suppresses calling, with the request held fixed in each arm (\autoref{sec:vector-state}). Changing this intermediate state therefore redirects the eventual decision before the lens gap crosses 1.0 and before the strongest MLP call-token writes. Together, these results place call-state formation between early access to the request distinction and the late output writes: the request contrast is already encoded when the formation-window computation produces a state that can regulate the subsequent decision.

\begin{table}[h]
  \centering
  \small
  \caption{Held-out early-signal diagnostics and layer-wise gap accumulation on the 200 held-out pairs (single-sample TransformerLens protocol with standard single RMSNorm).}
  \label{tab:early-signal-triplet}
  \begin{tabular}{ll}
    \toprule
    Diagnostic & Result \\
    \midrule
    Probe AUC $> 0.9$ first layer & L1 \\
    Logit-lens gap $> 1$ first layer & L27 \\
    DLA top-3 layers (MLP write) & L35, L34, L33 \\
    Final gap mean (single RMSNorm, 95\% CI) & 7.093 [6.687, 7.498] \\
    First 10\% / 25\% / 50\% of final gap & L25 / L29 / L29 \\
    Largest increments & L29, L33, L24, L27, L32 \\
    \bottomrule
  \end{tabular}
\end{table}

\subsection{Analysis-Active Feature Writes Suppress the Call Tendency}
\label{appendix:transcoder}
The main text defines the signed feature contribution $\kappa$ and the two activation-side contribution masses. To interpret the writes forming the L24 state, we select the 80 largest positive and 80 largest negative training-$\kappa$ candidates per layer across L20--L23. This gives 640 distinct layer--feature candidates from the 655,360 features in the formation window. Selection and family assignment use the 300 training pairs; contribution magnitudes are evaluated on the 200 held-out pairs.

Family assignment combines decoder-token cues with the request contexts that activate each feature. We inspect the 20 highest- and 20 lowest-projection tokens under each decoder and the five highest-activating training examples, retaining their three most frequent request verbs. Fixed lexical rules score non-execution, execution, discourse, and schema cues together with the activation-side/write-direction pattern. A candidate receives the highest-scoring family if its score reaches 4.0, or 3.0 for \texttt{schema\_boundary\_formatting}. This assigns 426 candidates; all 640 candidates and the remaining dictionary features retain their contributions in the full-feature $K$ accounting.

\begin{table}[h]
\centering
\small
  \caption{Formation-window feature families selected on training pairs.
Side is the majority activation side on training pairs. Counts cover
426 of 640 candidates; 214 remain unlabelled. Mean and sum use held-out $|\kappa|$.}
\label{tab:appendix-transcoder-families}
\begin{tabularx}{\linewidth}{lcrrrX}
\toprule
Family & Side & $N_f$ & Mean & Sum & Interpretation \\
\midrule
  \texttt{analysis\_non\_execution} & corrupt & 115 & 0.0633 & 7.2787 & Analysis requests and non-necessity semantics. \\
  \texttt{execution\_request} & clean & 292 & 0.0156 & 4.5634 & Execution requests and action verbs. \\
  \texttt{analysis\_plain\_discourse} & corrupt & 15 & 0.0020 & 0.0294 & General analysis discourse. \\
  \texttt{schema\_boundary\_formatting} & corrupt & 4 & 0.0466 & 0.1862 & Code and format boundaries. \\
\bottomrule
\end{tabularx}
\end{table}

The family comparison asks which request-associated writes account for the analysis-side contribution mass. Although \texttt{analysis\_non\_execution} contains fewer labelled features than \texttt{execution\_request}, its mean $|\kappa|$ is about four times larger, giving it the largest family total (Table~\ref{tab:appendix-transcoder-families}). The concentration therefore comes from stronger contributions per feature. For example, L23 F117270 projects strongly onto tokens such as \emph{nowhere}, \emph{none}, \emph{neither}, and \emph{nothing}, and its highest-activating training examples use \emph{Discuss} and \emph{Explore}. Together with the small contribution of \texttt{analysis\_plain\_discourse}, these cues associate the dominant labelled writes with non-execution.

The decoder directions explain how these analysis-side contributions enlarge the paired gap. On the training pairs, 102 of the 115 \texttt{analysis\_non\_execution} features have higher analysis-side activation and a decoder projection onto $\hat{\mu}_\Delta$ that is negative. Greater activation therefore makes the analysis-state write more negative; subtracting that analysis state from the execution state gives a positive contribution to the gap. Combined with the high calling baseline under neutral requests (Appendix~\ref{appendix:scaffold-controls}), these signs support active suppression of a prepared call tendency on the analysis side. The feature contexts identify the requests recruiting this computation, and the signed writes explain its effect on the call state.

We next test whether the features identified by this signed decomposition exert behavioral influence under the same analysis requests. If their writes implement suppression, removing them should move the L24 state toward execution and increase calling. We zero the prediction-position writes of the five analysis-higher features with the largest training $|\kappa|$ whose decoder directions oppose $\hat{\mu}_\Delta$. These five features constitute less than 0.001\% of the formation-window dictionary, yet their removal moves the state along $+\hat{\mu}_\Delta$ and raises calling from 3.0\% to 25.5\% (Table~\ref{tab:appendix-suppressor-ablation}). The state and behavioral changes follow the predicted direction, connecting the decomposition to causal suppression; the layer-matched comparison ties the effect to the selected writes.

\begin{table}[h]
\centering
\small
\caption{Suppressor-write ablation on 200 feature-evaluation pairs, with 6 baseline analysis-side calls (3.0\%). $\Delta_\parallel$ is the change in the L24 projection along $\hat{\mu}_\Delta$; gap closure is relative to the baseline execution--analysis gap. Call is the post-intervention call rate; strict recovery counts baseline text decisions changed to calls, divided by all 200 pairs.}
\label{tab:appendix-suppressor-ablation}
\begin{tabular}{lrrrr}
\toprule
Ablated writes & $\Delta_\parallel$ & Gap closed (\%) & Call (\%) & Strict recovery (\%) \\
\midrule
Five training-ranked suppressors & $+5.30$ & 8.82 & 25.5 & 22.5 \\
Layer-matched random suppressors & $+0.14$ & 0.23 & 2.5 & 0.5 \\
\bottomrule
\end{tabular}
\vspace{-4pt}
\end{table}

\subsection{Linking Formation-Window MLP Writes to the Call State}
\label{appendix:formation:mediation}
The feature ablation identifies individual writes with a suppressive effect. We next ask whether the formation window's aggregate MLP effect on calling passes through the L24 call-direction coordinate. We replace all L20--L23 prediction-position MLP outputs with their paired opposite-side outputs, then restore only the original projection along $\hat{\mu}_\Delta$ at L24. The orthogonal component retains its swapped value, so the rescue tests whether restoring that single coordinate can recover the original call decision after a much broader perturbation.

On execution prompts, swapping the MLP outputs reduces calling to 93.0\%. Restoring only the original call-direction projection recovers 100.0\% calling, compared with 58.0\% after orthogonal restoration (Table~\ref{tab:appendix-formation-mediation}). The execution decision therefore recovers while the state's orthogonal component retains its swapped value. In the reverse arm, restoring the analysis-side projection lowers calling from 95.0\% after the swap to 0.0\%. The selective effects in both directions connect formation-window MLP computation to tool calling.

\begin{table}[h]
\centering
\small
\caption{Formation-window mediation on the 200 feature-evaluation pairs.
Each arm computes $g_l$ against the untouched paired opposite-side state.
Call rate is the post-intervention top-1 rate.}
\label{tab:appendix-formation-mediation}
\begin{tabular}{lrrrr}
\toprule
& \multicolumn{2}{c}{Execution prompts} & \multicolumn{2}{c}{Analysis prompts} \\
\cmidrule(lr){2-3}\cmidrule(lr){4-5}
Condition & $g_l$ & Call (\%) & $g_l$ & Call (\%) \\
\midrule
Baseline & 60.06 & 100.0 & 60.06 & 3.0 \\
MLP-output swap & 15.78 & 93.0 & 15.09 & 95.0 \\
Swap + $\hat{\mu}_\Delta$ restoration & 60.06 & 100.0 & 60.06 & 0.0 \\
Swap + orthogonal restoration & 15.78 & 58.0 & 15.09 & 99.5 \\
Swap + full-state restoration & 60.06 & 100.0 & 60.06 & 3.0 \\
\bottomrule
\end{tabular}
\vspace{-4pt}
\end{table}

\section{Reading Out the Tool-Call Vector}
\label{appendix:readout}
Downstream attention heads and MLPs translate the L24 call state into support for the first output token. We examine how components in L25--L35 respond to this state, use scaffold information, and contribute to call initiation.

\subsection{Attention Heads and MLPs Contribute at Multiple Readout Stages}
For each downstream attention head and MLP block, we compare direct output support with the effect of an intervention. The clean-minus-corrupt DLA delta measures the difference in the component's direct contribution to the $\texttt{<tool\_call>}$ logit between execution and analysis prompts. Single-component patching replaces an analysis-side output with its paired execution-side output and measures the resulting logit shift after downstream computation resumes. Strict flip is the fraction of all evaluated pairs whose baseline analysis response is text and whose patched response opens a call. Reverse patching measures the logit drop on execution prompts. Comparing these measurements relates each component's direct write to its influence through the remaining computation.

\begin{table}[h]
  \centering
  \small
  \caption{Selected attention heads in the L25--L35 component sweep (baseline corrupt call rate 3.6\%), also shown in Figure~\ref{fig:readout}A. DLA delta is clean minus corrupt for the \texttt{<tool\_call>} logit. Recovery shift and strict flip measure replacement of the corrupt-side head output with its output.}
  \label{tab:appendix-readout-heads}
  \begin{tabular}{llrrrr}
    \toprule
    Head & stage & DLA $\Delta$ & recover shift & corrupt top-1 & strict flip \\
    \midrule
    L25H6  & earlier & 0.06 & 1.28 & 40.4\% & 36.9\% \\
    L28H3  & earlier & 0.26 & 1.38 & 36.1\% & 32.5\% \\
    L26H14 & earlier & 0.04 & 0.91 & 24.3\% & 20.7\% \\
    L29H9  & earlier & 4.15 & 1.34 & 21.3\% & 17.8\% \\
    L33H29 & late & 7.91 & 0.62 & 8.5\% & 4.9\% \\
    L33H11 & late & 3.32 & -0.39 & 2.4\% & 0.0\% \\
    \bottomrule
  \end{tabular}
\end{table}

The contrast between DLA and patching reveals complementary roles in call readout. L25H6 and L28H3 have small DLA differences but substantial patching recovery, showing that their effects extend through the remaining computation (Table~\ref{tab:appendix-readout-heads}). The L33 heads show larger differences in direct output support. These results distinguish the influence of earlier components on downstream processing from the direct call-token support supplied by later writers.

The MLP results place effective readout contributions at several downstream layers. MLP34 gives the highest recovery in the component sweep, while MLP25 and MLP27 also produce large effects (Table~\ref{tab:appendix-readout-mlps}). This distribution connects the L24 state to calling through both earlier and later MLP computation. We focus the feature analysis on MLP34 because its strong patching effect and late call-directed write provide a useful site for resolving how the state is expressed in the output format.

\begin{table}[t]
  \centering
  \small
  \caption{Top MLP blocks in the L25--L35 component sweep (baseline corrupt call rate 3.6\%). Recovery replaces a corrupt-side MLP output with its paired clean-side output; the reverse replacement on clean prompts gives the drop measurement.}
  \label{tab:appendix-readout-mlps}
  \begin{tabular}{lrrrr}
    \toprule
    MLP & recover shift & drop shift & corrupt top-1 & strict flip \\
    \midrule
    L34 & 1.74 & 2.32 & 53.3\% & 49.7\% \\
    L27 & 1.32 & 0.82 & 52.3\% & 48.7\% \\
    L25 & 1.70 & 0.96 & 44.8\% & 41.2\% \\
    L35 & 0.84 & 1.27 & 31.8\% & 28.2\% \\
    L31 & 0.90 & 0.90 & 28.6\% & 25.0\% \\
    L29 & 0.38 & 1.58 & 26.6\% & 23.1\% \\
    \bottomrule
  \end{tabular}
\end{table}

\subsection{The Call State Regulates Downstream Call-Token Writes}
L29H9, L33H11, and L33H29 attend more strongly to the tool-format template on execution prompts, although the scaffold is identical within each held-out pair (Figure~\ref{fig:readout}B). We ask whether changing the internal call state can recruit these heads' call-directed writes while the request and scaffold remain fixed. Adding $\mu_\Delta$ at L24 on analysis prompts brings all three heads' call-token DLA close to their execution-side values (Table~\ref{tab:appendix-readout-vector-response}). The same input therefore supports stronger writes toward the call opening when the intermediate state is shifted toward calling.

To test the readout in the suppression direction, we subtract $\mu_\Delta$ at L24 on execution prompts in a separate attribution control. Total L30--L35 call-token DLA falls by 35.45\%, with the largest absolute drops in L33--L35. The weaker output writes occur downstream of the intervention while the prompt remains fixed. Together with the head-level response to addition, this shows that changing the intermediate call state regulates how strongly subsequent components support the call-opening token.

\begin{table}[h]
\centering
\small
\caption{Attention-head response to L24 vector addition on the 200 balanced
Qwen3-8B held-out pairs. Entries are mean DLA for the call-opening token.}
\label{tab:appendix-readout-vector-response}
\begin{tabular}{lrrr}
\toprule
Head & Clean & Corrupt & Corrupt $+\mu_\Delta$ \\
\midrule
L29H9 & 6.99 & 1.43 & 6.91 \\
L33H11 & 5.60 & 1.46 & 5.30 \\
L33H29 & 24.58 & 11.13 & 24.43 \\
\bottomrule
\end{tabular}
\end{table}

\subsection{MLP34 Feature F109925 Supports the Tool-Call Opening}
F109925 offers a concrete link between the residual state and the output format. Its max-activating contexts highlight tool-schema boundaries, and its decoder writes toward the call-opening token. Adding $\mu_\Delta$ to analysis prompts restores feature activation toward its execution-side level, increasing call-directed output through that fixed decoder (Table~\ref{tab:appendix-structural-readout}). The state intervention therefore recruits a feature capable of supporting the call opening, while the activating contexts provide a structural interpretation of its write.

\begin{table}[h]
\centering
\small
\caption{L34 F109925 on the 200 Qwen3-8B feature-evaluation held-out pairs.
Projected write is activation multiplied by the feature decoder's
projection onto the call-token unembedding direction.}
\label{tab:appendix-structural-readout}
\begin{tabular}{lrrr}
\toprule
Measurement & Clean & Corrupt & Corrupt $+\mu_\Delta$ \\
\midrule
Mean activation & 124.43 & 67.63 & 126.00 \\
Projected call-token write & 6.17 & 3.35 & 6.25 \\
\bottomrule
\end{tabular}
\end{table}

To test whether this structural write affects the decision, we replace F109925's analysis-side activation at the prediction position with its paired execution-side value through the feature decoder, using the native normalized MLP input. The replacement yields 37.0\% calling, approaching the full MLP34 patch's 41.0\%, with 6.5\% calling for the fixed non-structural control F91365 (Table~\ref{tab:appendix-feature-readout-controls}). Changing this single feature supplies enough call-directed support to cross the decision boundary on many prompts. Together with its activation response to $\mu_\Delta$, the behavioral effect identifies a structural write through which the call state promotes the call opening.

\begin{table}[h]
\centering
\small
\caption{L34 feature replacement and controls on the same 200 held-out pairs.
All interventions act at the prediction position.}
\label{tab:appendix-feature-readout-controls}
\begin{tabular}{lr}
\toprule
Condition & Call (\%) \\
\midrule
Corrupt $+\mu_\Delta$ at L24 & 100.0 \\
Replace F109925 & 37.0 \\
Replace F91365 & 6.5 \\
Patch full MLP34 & 41.0 \\
\bottomrule
\end{tabular}
\end{table}

\section{Cross-Model Timing and Scaffold Attention}
\label{appendix:cross-scale}
These diagnostics relate the model-specific call states in \autoref{tab:crossscale} to the timing of request encoding and the use of native scaffold information.

\subsection{Early Request Encoding and Late Call-Token Writes Across Models}
\label{appendix:cross-scale:triplet}

We measure request separability and call-token expression across all seven models to trace their progression toward the output decision. The held-out diagnostic cohorts contain 200 pairs for Qwen3-8B, 240 for Granite-3.3-8B, and 300 for each of the other five models (Table~\ref{tab:appendix-crossscale-triplet}).
For each model, we measure:
(i) the linear probe AUC along the residual stream to detect when clean and corrupt representations become linearly separable;
(ii) the logit-lens gap $\Delta \mathrm{logit}(l) = \operatorname{logit}_{\mathrm{clean}}(l) - \operatorname{logit}_{\mathrm{corrupt}}(l)$ projected onto the model's native tool-call opening token unembedding direction; and
(iii) the layer-wise Direct Logit Attribution (DLA) $\Delta \mathrm{DLA}(l) = \mathrm{DLA}_{\mathrm{clean}}(l) - \mathrm{DLA}_{\mathrm{corrupt}}(l)$ of each block's net residual write, combining attention and MLP outputs, to measure direct support for the call token.

\begin{figure}[p]
  \centering
  \includegraphics[width=\linewidth,height=0.84\textheight,keepaspectratio]{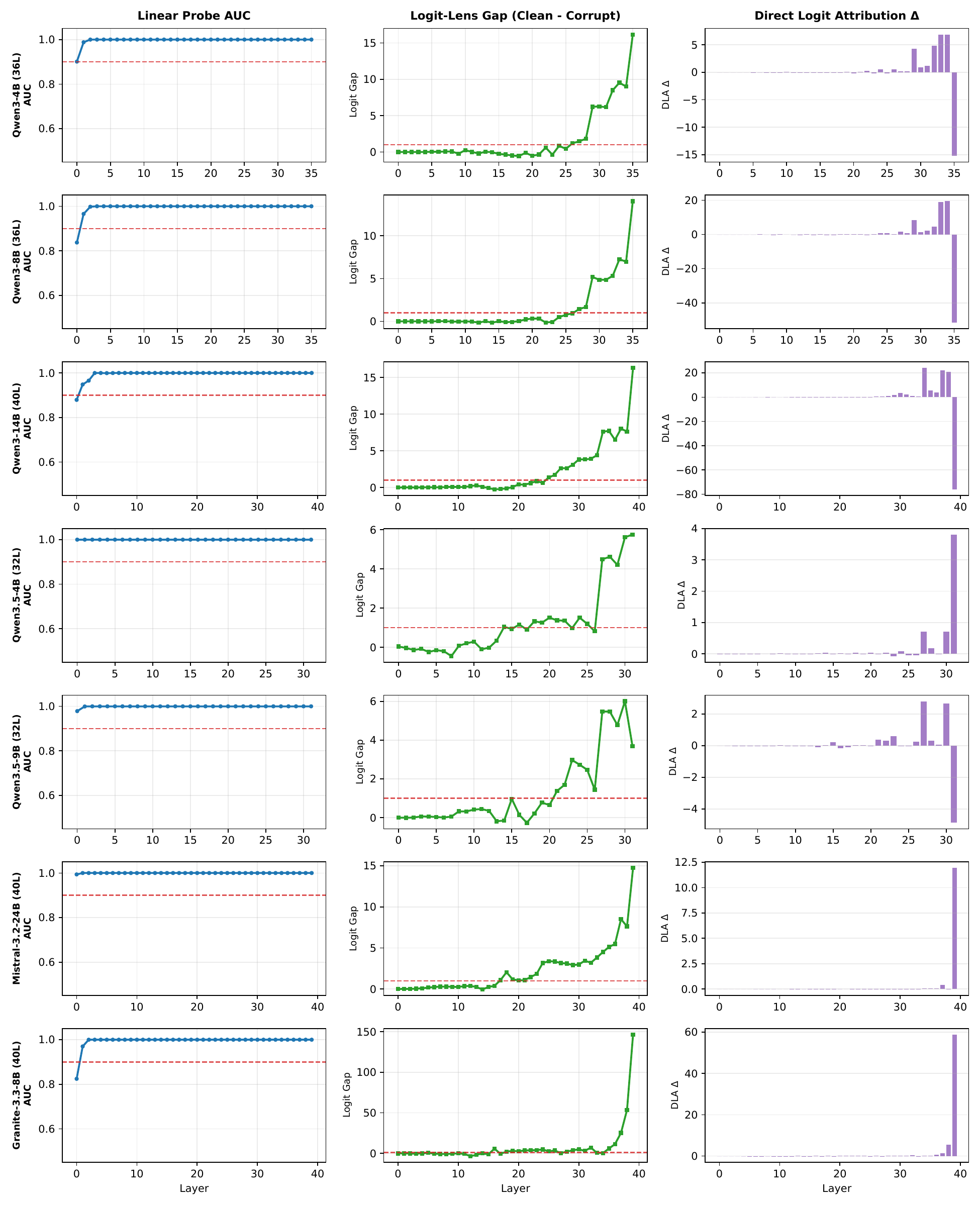}
  \caption{Triplet diagnostics across all seven models, arranged with one model
  per row and three diagnostics per row. From left to right, each row shows the
  linear probe AUC, the logit-lens clean-minus-corrupt gap, and the layer-wise DLA sweep.
  Evaluation cohorts contain 200 pairs for Qwen3-8B, 240 for Granite-3.3-8B,
  and 300 for each of the other five models. Across all seven models, probe
  separability appears at L0--L1, while the largest DLA writes occur in upper layers.}
  \label{fig:appendix-crossscale-triplet-curves}
\end{figure}

Table~\ref{tab:appendix-crossscale-triplet} summarizes the timing landmarks across all seven models, and Figure~\ref{fig:appendix-crossscale-triplet-curves} displays the full layer-by-layer curves.

\begin{table}[h]
\centering
\small
\caption{Early-signal and timing diagnostics across all seven models on held-out prompt pairs.
$L^*$ is the causal intervention layer from Table~\ref{tab:crossscale};
Probe AUC $> 0.9$ is the first layer where the linear probe achieves $>0.9$ AUC;
Gap $> 1.0$ is the first layer where the logit-lens clean-minus-corrupt gap exceeds 1.0;
Top-3 DLA reports the blocks with the largest signed positive net direct logit attribution differences ($\Delta \mathrm{DLA} > 0$), combining attention and MLP writes. $L^*$ denotes a block input; probe and lens layer labels denote block-output residuals, and DLA labels denote block writes.}
\label{tab:appendix-crossscale-triplet}
\resizebox{\linewidth}{!}{%
\begin{tabular}{lcccccc}
\toprule
Model & $L_\mathrm{tot}$ & $L^*$ & $N$ & Probe AUC $> 0.9$ & Gap $> 1.0$ & Top-3 DLA layers \\
\midrule
Qwen3-4B & 36 & L26 & 300 & L0 & L26 & L34, L33, L32 \\
Qwen3-8B & 36 & L24 & 200 & L1 & L27 & L34, L33, L29 \\
Qwen3-14B & 40 & L34 & 300 & L1 & L25 & L34, L37, L38 \\
Qwen3.5-4B & 32 & L31 & 300 & L0 & L14 & L31, L30, L27 \\
Qwen3.5-9B & 32 & L31 & 300 & L0 & L21 & L27, L30, L23 \\
Mistral-Small-3.2-24B & 40 & L25 & 300 & L0 & L17 & L39, L37, L34 \\
Granite-3.3-8B & 40 & L35 & 240 & L1 & L16 & L39, L38, L37 \\
\bottomrule
\end{tabular}
\hspace*{-0.2em}}
\end{table}

Across all seven models, request information is recoverable at L0--L1, while the strongest direct call-token writes occur in upper layers. This shared separation places request encoding early enough to guide substantial subsequent computation, with the largest direct expression of the contrast concentrated near the output. Together with the interventions in \autoref{tab:crossscale}, these diagnostics connect early request information to a causally controllable call state and late output support across architectures.

The relation between vocabulary alignment and the effective intervention state varies across models. Qwen3-8B supports call control at L24 before its lens gap crosses 1.0 at L27; Granite crosses that threshold at L16 and supports intervention at L35 (Table~\ref{tab:appendix-crossscale-triplet}). These measurements expose complementary aspects of the computation: the lens tracks developing output preference, and the intervention identifies a state through which the remaining network can be directed toward a different call decision. Their model-specific timing accompanies the early-encoding and late-writing pattern.

\subsection{Execution Requests Increase Attention to Native Tool Scaffolds}
\label{appendix:cross-scale:readers}
Native scaffold attention tests how the request contrast changes the use of the tool interface already present in the prompt. Max Attn in \autoref{tab:crossscale} averages each measured head's execution-minus-analysis attention shift over native held-out pairs, then takes the largest head-level mean. The target region is the invocation-format template for Qwen-family models and the tool-schema region for Mistral and Granite. Qwen3-8B uses its three reported scaffold readers; Qwen3.5 measurements use full-attention layers.

The positive shifts connect execution requests to increased attention to call-relevant scaffold content. Because the scaffold stays fixed within a pair, the change reflects how the request recruits the available interface information. This extends the Qwen3-8B attention analysis to each model's native serialization: format instructions and tool definitions provide the scaffold information through which a call is expressed.

\section{Circuit Discovery Diagnostics}
\label{appendix:circuit}
We use ACDC and EAP-IG to examine which component sets and interactions sustain the call contrast. Their replay and removal controls relate the compact residual-state intervention to the distributed computation supporting the decision.

\subsection{ACDC: Sparse-Core Replay and Removal Across Qwen3 Scales}
\label{appendix:circuit:acdc}
ACDC~\citep{conmy2023acdc} prunes candidate component connections according to their effect under activation patching. We evaluate the resulting sparse forward cores across Qwen3 scales, including the additional Qwen3-1.7B diagnostic.

We compare a core's ability to carry the task signal under replay with its contribution in the intact network (Table~\ref{tab:appendix-acdc-faithfulness}). Sufficiency patches the selected clean component outputs into the analysis run; removal replaces those outputs in the clean run with their paired analysis-side values. Let $s_{\mathrm{clean}}$, $s_{\mathrm{analysis}}$, $s_{\mathrm{replay}}$, and $s_{\mathrm{removal}}$ denote the corresponding next-token task scores. For each pair, sufficiency is $(s_{\mathrm{replay}}-s_{\mathrm{analysis}})/(s_{\mathrm{clean}}-s_{\mathrm{analysis}})$, and the removal effect is $(s_{\mathrm{clean}}-s_{\mathrm{removal}})/(s_{\mathrm{clean}}-s_{\mathrm{analysis}})$. We report mean ratios and compare replay with size-matched random component sets.

\begin{table}[h]
  \centering
  \small
  \caption{Sparse forward-core replay across Qwen3 scales.
  Core nodes and edges are taken from the aggregated forward sparse graph.
  Suff.\ is the mean normalized score recovery under core replay; Removal is the
  mean normalized score reduction under clean-side replacement of the core.
  Rand.\ Suff.\ is the mean recovery for size-matched random component sets,
  and Gap is the discovered core's mean advantage over that baseline.}
  \label{tab:appendix-acdc-faithfulness}
  \begin{tabular}{lcccccc}
  \toprule
  Model & Core nodes & Core edges & Suff. & Removal & Rand.\ Suff. & Gap \\
  \midrule
  Qwen3-1.7B & 18 & 54 & 0.9891 & 0.9824 & 0.8812 & 0.1078 \\
  Qwen3-4B & 18 & 69 & 0.9979 & 0.8319 & 0.8712 & 0.1266 \\
  Qwen3-8B & 18 & 63 & 1.0000 & 0.5975 & 0.9768 & 0.0232 \\
  Qwen3-14B & 20 & 86 & 0.9999 & 0.4077 & 0.9554 & 0.0446 \\
  \bottomrule
\end{tabular}
\end{table}

The cores recover scores close to the clean baseline under replay at every tested scale. At 8B and 14B, the size-matched random sets also recover much of that score, so the replayed signal can pass through several component sets. Clean-side replacement reveals the selected cores' influence within the full computation: the normalized score reductions are 0.5975 and 0.4077, respectively (Table~\ref{tab:appendix-acdc-faithfulness}). The combination links effective sparse replay to contributions spread across the intact network, motivating the shared residual state as a way to trace their collective effect on calling.

\subsection{EAP-IG: Recovery of the Call Contrast Across Edge Budgets}
\label{appendix:circuit:eap}
To examine how recovery changes as more of the computation is retained, we rank edges with EAP-IG~\citep{hanna2024faithfulness}. Integrated gradients estimate the effect of moving edge activations between paired analysis and execution runs; retaining top-ranked edges or greedily connected edges gives circuits at different budgets. Gradient-aware variants further refine this attribution procedure~\citep{zhang2025eapgp}. We evaluate each retained graph by its recovery of the call-related logit-difference score, normalized between the analysis and execution baselines.

\begin{figure}[h]
  \centering
  \includegraphics[width=0.78\linewidth]{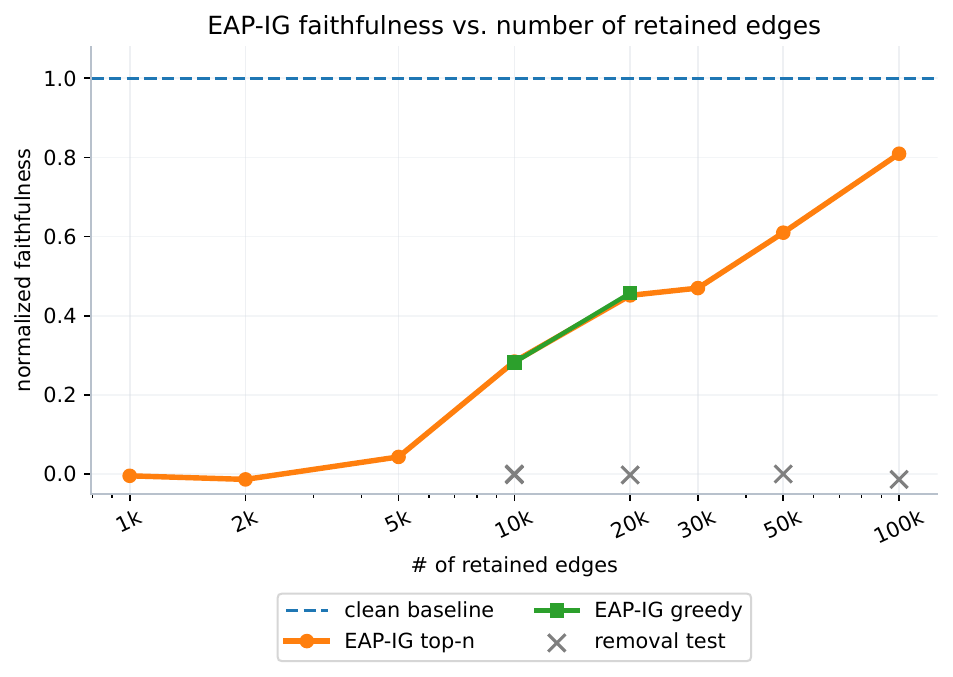}
  \caption{EAP-IG top-$k$ faithfulness curve for the tool-call task, normalized
  between the corrupt and clean baselines. The curve relates retained edge
  budget to recovery of the \texttt{<tool\_call>} signal.}
  \label{fig:appendix-eap-ig-faithfulness}
\end{figure}

\begin{table}[h]
  \centering
  \small
  \caption{EAP-IG top-$k$ sweep on Qwen3-8B using 100 equal-length clean/corrupt
  prompt pairs. Faith.\ normalizes sufficiency between the corrupt and clean
  baselines. Suff.\ is the mean logit-difference score under corrupt-side
  replay with only the retained edges active. Nodes counts the retained
  component set induced by the edges; core overlap counts shared nodes with
  the 18-node forward core.}
  \label{tab:appendix-eap-ig-topk}
  \begin{tabular}{rrrrc}
  \toprule
  Retained edges & Faith. & Suff. logit diff & Nodes & Core overlap \\
  \midrule
  1,000 & -0.005 & -13.12 & 139 & 17/18 \\
  2,000 & -0.014 & -13.38 & 229 & 18/18 \\
  5,000 & 0.043 & -11.81 & 417 & 18/18 \\
  10,000 & 0.284 & -5.19 & 553 & 18/18 \\
  20,000 & 0.451 & -0.59 & 708 & 18/18 \\
  30,000 & 0.470 & -0.08 & 805 & 18/18 \\
  50,000 & 0.610 & 3.77 & 937 & 18/18 \\
  100,000 & 0.809 & 9.25 & 1,075 & 18/18 \\
  \bottomrule
\end{tabular}
\end{table}

The full 18-node forward core is already represented at 2,000 retained edges, before substantial task recovery appears. As the retained computation expands in both nodes and edges, faithfulness rises to 0.809 at 100,000 edges (Table~\ref{tab:appendix-eap-ig-topk}). Thus the contribution of the core develops through a broader set of components and interactions. Clean-side removal of the retained edges also strongly reduces the task score (Figure~\ref{fig:appendix-eap-ig-faithfulness}), linking that expanded computation to the call contrast. These replay and removal results complement the vector intervention by locating distributed support for the shared call-controlling state.

\section{Related Work}
\label{appendix:relatedwork}

This section relates the tool-call decision to three lines of work: agentic tool-use systems, mechanistic analysis at the representation, circuit, and feature levels, and behavior-mediating residual directions.

\subsection{Tool-Calling and Agentic Systems in Large Language Models}
\label{appendix:relatedwork:tools}

\paragraph{Training and prompting for tool use.}
Enabling large language models to interact with external environments has
progressed across several training and prompting paradigms. Early methods
introduced tool-calling capabilities by fine-tuning models on synthetically generated
or bootstrap-filtered tool execution trajectories \citep{schick2023toolformer, patil2023gorilla, qin2023toolllm, li2023apibank}.
Concurrent prompting frameworks established multi-step interaction protocols,
most notably ReAct \citep{yao2023react}, where models alternate between reasoning traces
and environment actions. Modern open-weight foundation models, such as the Qwen family
\citep{yang2025qwen3}, IBM Granite, and Mistral-Small, incorporate dedicated agentic
pre-training and instruction tuning, allowing models to natively output structured
special tokens (e.g., \texttt{<tool\_call>}) and parse tool response blocks without
external wrapper scripts.

\paragraph{Evaluation benchmarks and execution environments.}
Evaluation suites span single-turn API selection, stateful tool use, and agentic
instruction following. MetaTool evaluates tool awareness and selection \citep{huang2024metatool},
while BFCL covers function calling and stateful multi-step agentic evaluation \citep{patil2025bfcl}.
$\tau$-Bench \citep{yao2024taubench} and $\tau^2$-Bench \citep{barres2025tau2bench}
evaluate sequential interaction in simulated domains, while AgentIF focuses on instruction
following under long and complex agentic constraints \citep{qi2025agentif}.
Code-generation benchmarks, including MBPP \citep{austin2021mbpp}, APPS \citep{hendrycks2021apps},
HumanEval \citep{chen2021codex}, and CodeContests \citep{li2022alphacode}, provide
naturally structured environments where a tool interface (e.g., \texttt{write\_file} or an interpreter)
is embedded directly into complex problem specifications. Robustness evaluations further reveal
that tool-calling models can be sensitive to superficial prompt variations, format drift,
and adversarial perturbations \citep{sclar2024formatspread, rabinovich2025robustfc, xia2025safetoolbench}.

\paragraph{Behavioral outcomes versus internal mechanisms.}
Tool-use benchmarks measure tool selection, argument construction, action accuracy, and task completion. Our analysis examines the internal computation leading to the first-token choice that initiates a tool call, connecting a specific stage of these broader behaviors to a causally controllable residual state.

\subsection{Mechanistic Interpretability: Representations, Circuits, and Features}
\label{appendix:relatedwork:interpretability}

Mechanistic interpretability aims to reverse-engineer neural networks into human-understandable
computational graphs and representational states. We contextualize our methodology across
three distinct levels of granularity: the representation level, the circuit level, and the feature level.

\paragraph{Representation-level localization and causal interventions.}
Representation-level methods connect internal states to behavior through complementary measurements. Linear probes test recoverability across layers~\citep{alain2016linearprobes, hewitt2019controlprobes, belinkov2022probing}, while vocabulary projections such as the logit lens and tuned lens track developing token preferences~\citep{belrose2023tunedlens}. DLA resolves direct output-logit contributions~\citep{elhage2021framework}, and activation patching and interchange interventions test the effects of replacing internal states~\citep{meng2022locating, heimersheim2024patching}. Their combination supports the distinction between request information, causal call control, and downstream output support used in our analysis.

\paragraph{Circuit-level analysis of tool-call computation.}
Circuit analysis identifies component interactions that reproduce a task computation \citep{wang2022ioi, conmy2023acdc, syed2023attribution}. Automated methods such as ACDC and Edge Attribution Patching map candidate paths through attention heads and MLPs; gradient-aware discovery \citep{zhang2025eapgp} and self-influence analysis \citep{zhang2025mechanistic} further examine how these interactions support behavior. Tool-call initiation integrates information from request wording, task content, tool definitions, and invocation templates. Circuit diagnostics describe how this information contributes across components (Appendix~\ref{appendix:circuit}), while the residual-state intervention tests their shared decision-level effect. Transcoder decomposition then assigns signed contributions to the features forming the state, connecting component interactions to a causally controllable representation.

\paragraph{Feature-level decomposition via Transcoders.}
Sparse dictionary methods resolve MLP computation into feature activations and writes~\citep{geva2023dissecting}. SAEs reconstruct internal activations~\citep{cunningham2023sae, gao2024scaling, lawson2024multilayersae, marks2024sparsefeaturecircuits}, and Transcoders model the MLP input-to-output mapping as sparse activations multiplied by decoder directions~\citep{dunefsky2024transcoders}. This factorization connects the contexts recruiting a feature to the direction in which it changes the residual state. It supplies the signed write analysis linking request-sensitive formation features and structural readout features to the call state. Work on planning features across scales~\citep{hanna2026latentplanningemergesscale} further motivates comparing this organization across model families.

\subsection{Behavior-Mediating Directions and Steering Vectors}
\label{appendix:relatedwork:steering}

\paragraph{Representation engineering and activation steering.}
A growing body of literature demonstrates that complex model behaviors can be mediated and
controlled by compact directions in residual-stream space. Representation engineering \citep{zou2023repe}
extracts behavioral concepts via contrastive pairs and injects them to control model attributes.
Related steering techniques, including activation addition \citep{turner2023activation}, Contrastive
Activation Addition (CAA) \citep{panickssery2023caa}, and Inference-Time Intervention (ITI) \citep{li2023iti},
show that targeted vector additions can alter truthfulness, sycophancy, or stylistic persona without
retraining model weights. Recent frameworks further refine steering precision and stability:
\citet{jiang2026global} introduce Global Evolutionary Steering to enforce cross-layer consistency,
\citet{yu2026pixel} develop PIXEL for adaptive position-wise injection with subspace calibration,
and \citet{yu2026multiadapter} propose MARI to calibrate representation interventions across multi-adapter
architectures. Similarly, function vectors \citep{todd2024functionvectors} demonstrate that
in-context task execution routines compress into localized vectors that can trigger specific behaviors
when injected into zero-shot contexts.

\paragraph{The refusal direction and default-and-suppression dynamics.}
The closest conceptual precedent to our findings is the refusal direction described by \citet{arditi2024refusal}.
They show that refusal in safety-aligned language models is mediated by a single residual-stream
direction: ablating this direction suppresses refusal, while adding it can induce refusal on harmless
requests. Our analysis reveals an analogous functional organization in tool calling:
the prompt scaffold establishes a strong call prior on task-bearing inputs,
while analysis requests activate suppressor features that write against $\hat{\mu}_\Delta$ to
inhibit call initiation (\autoref{sec:formation}).

The transfer experiments connect this mechanistic account to the reuse of steering directions. A direction fitted in the controlled request contrast remains effective under new tasks, interfaces, and request forms: cross-domain tests calibrate its magnitude on separate target-domain training pairs, while the $\tau^2$-Bench and verb-free tests use fixed gains of the original vector. This connects the state-level intervention to a reusable action-initiation signal.

\section{Implementation and Reproducibility}
\label{appendix:implementation}
The implementation uses PyTorch and Hugging Face Transformers. Frozen prompt
manifests record the construction models, data splits, and file hashes.
Model-specific templates preserve the native tool-call serialization.

\subsection{Evaluation Protocol}
\label{appendix:implementation:compute}

Vector fitting uses training pairs only. Evaluation applies interventions at the final input-token position and measures the logits of the first output token. Layer indices are zero-based, and $L^*$ denotes the block-input residual. The localization sweep compares the request-verb and prediction positions across decoder layers. The Qwen3-8B attention and feature-response analyses cover L25--L35 on the 200 balanced held-out pairs. Qwen3.5-4B formation-feature mass is evaluated over L28--L30. Cross-model attention measurements use each model's native held-out inputs and native call-opening token. The main paired experiments use a fixed 300/200 training/held-out split and the fixed block-input intervention layers. Additional appendix controls use the cohorts specified in their subsections.

Cross-domain transfer keeps the fitted direction fixed and calibrates its magnitude on target-domain training pairs. For Qwen3, Qwen3.5, Mistral, and Granite, intervention rates are evaluated on separate held-out pairs. The $\tau^2$-Bench and verb-free tests apply fixed gains to the original fitted vector.

\subsection{Code, Data, and Intermediate Artifacts}
\label{appendix:implementation:release}

The release comprises experiment code, the model-specific paired datasets,
and intermediate artifacts needed to reproduce the conclusions, including
Transcoder checkpoints. The repositories are:
\begin{itemize}
  \item Code and experiment instructions:
  \url{https://github.com/XijieGo/MI4ToolCalling}.
  \item Paired datasets and manifests:
  \url{https://huggingface.co/datasets/XijieGong/MI4ToolCalling}.
  \item Newly trained Transcoders and checkpoint documentation:
  \url{https://huggingface.co/XijieGong/MI4ToolCalling}.
\end{itemize}
Repository README files and per-model manifests document the directory
layout and input formats; the Transcoder model card provides checkpoint
descriptions and loading instructions. Appendix~\ref{appendix:dataset}
documents the prompt construction and per-model splits.

Our original code, dataset-construction materials, and newly trained
Transcoders use Apache License 2.0. Source benchmark content and third-party
checkpoints retain their original licenses, summarized in
Table~\ref{tab:appendix-asset-licenses}.
The base models in the primary comparison are
\href{https://huggingface.co/Qwen/Qwen3-4B}{Qwen3-4B},
\href{https://huggingface.co/Qwen/Qwen3-8B}{Qwen3-8B},
\href{https://huggingface.co/Qwen/Qwen3-14B}{Qwen3-14B},
\href{https://huggingface.co/Qwen/Qwen3.5-4B}{Qwen3.5-4B},
\href{https://huggingface.co/Qwen/Qwen3.5-9B}{Qwen3.5-9B},
\href{https://huggingface.co/mistralai/Mistral-Small-3.2-24B-Instruct-2506}{Mistral-Small-3.2-24B-Instruct-2506},
and \href{https://huggingface.co/ibm-granite/granite-3.3-8b-instruct}{Granite-3.3-8B-Instruct}.
All seven use Apache License 2.0.
The Qwen3 Transcoders of \citet{hanna2026latentplanningemergesscale}
are available in the \texttt{mwhanna/qwen3-\{4b,8b,14b\}-transcoders}
repositories under the MIT license.
PyTorch uses the BSD 3-Clause license, and Hugging Face Transformers uses
Apache License 2.0.

\begin{table}[h]
  \centering
  \small
  \caption{Licenses of the source benchmarks. Links identify the original
  resources; the original papers are cited in the dataset and transfer sections.}
  \label{tab:appendix-asset-licenses}
  \begin{tabularx}{\linewidth}{lX}
    \toprule
    Source benchmark & License \\
    \midrule
    \href{https://github.com/hendrycks/apps}{APPS} & MIT \\
    \href{https://github.com/openai/human-eval}{HumanEval} & MIT \\
    \href{https://huggingface.co/datasets/google-research-datasets/mbpp}{MBPP} & CC BY 4.0 \\
    \href{https://github.com/google-deepmind/code_contests}{CodeContests} & CC BY 4.0 (data); Apache 2.0 (code) \\
    \href{https://github.com/sierra-research/tau2-bench}{\(\tau^2\)-Bench} & MIT \\
    \href{https://fever.ai/dataset/fever.html}{FEVER} & CC BY-SA 3.0 (data) \\
    \href{https://yale-lily.github.io/spider}{Spider 1.0} & CC BY-SA 4.0 (data); Apache 2.0 (code) \\
    \bottomrule
  \end{tabularx}
\end{table}

\subsection{Compute Resources}
\label{appendix:implementation:resources}

Training our Transcoders required approximately 800 NVIDIA B200 GPU-hours
in total, including debugging and production training. Using the provided
checkpoints, the other reported experiments can be reproduced on a single
NVIDIA RTX PRO 6000 Blackwell GPU with 96~GB of memory.

\subsection{LLM Usage}
\label{appendix:implementation:llm-usage}

GPT-5 was used to generate debugging examples for the data-construction pipeline. The models evaluated in the mechanistic experiments are documented in Appendix~\ref{appendix:implementation:release}; the additional Qwen3-1.7B replay diagnostic is specified in Appendix~\ref{appendix:circuit:acdc}.
\newpage
\section*{NeurIPS Paper Checklist}

\begin{enumerate}

\item {\bf Claims}
    \item[] Question: Do the main claims made in the abstract and introduction accurately reflect the paper's contributions and scope?
    \item[] Answer: \answerYes{}
    \item[] Justification: The abstract and introduction state the paper's main claims: a controlled first-token decision interface, a tool-call vector that transfers across settings, and an account of the vector's formation and downstream readout. Scope and boundary conditions are further clarified in Limitations.

\item {\bf Limitations}
    \item[] Question: Does the paper discuss the limitations of the work performed by the authors?
    \item[] Answer: \answerYes{}
    \item[] Justification: The paper includes a dedicated Limitations section that defines the scope as the binary call-or-no-call decision and identifies the unresolved fine-grained formation mechanism.

\item {\bf Theory assumptions and proofs}
    \item[] Question: For each theoretical result, does the paper provide the full set of assumptions and a complete (and correct) proof?
    \item[] Answer: \answerNA{}
    \item[] Justification: The paper is empirical and does not present formal theorems or proofs.

    \item {\bf Experimental result reproducibility}
    \item[] Question: Does the paper fully disclose all the information needed to reproduce the main experimental results of the paper to the extent that it affects the main claims and/or conclusions of the paper (regardless of whether the code and data are provided or not)?
    \item[] Answer: \answerYes{}
    \item[] Justification: The main text defines activation patching, the shared vector $\mu_\Delta$, and the sufficiency and necessity metrics. Appendix~\ref{appendix:dataset} describes the paired datasets, prompt formats, and per-model splits. Appendix~\ref{appendix:implementation} identifies the experiment code, datasets, and Transcoder checkpoints needed for reproduction. And we release the them.

\item {\bf Open access to data and code}
    \item[] Question: Does the paper provide open access to the data and code, with sufficient instructions to faithfully reproduce the main experimental results, as described in supplemental material?
    \item[] Answer: \answerYes{}
    \item[] Justification: The reproducibility package contains experiment code, paired datasets, and the intermediate artifacts needed for reproduction, including Transcoder checkpoints. Appendix~\ref{appendix:implementation:release} provides the GitHub and Hugging Face repository links and identifies the accompanying README files, input manifests, and checkpoint loading instructions.

\item {\bf Experimental setting/details}
    \item[] Question: Does the paper specify all the training and test details (e.g., data splits, hyperparameters, how they were chosen, type of optimizer) necessary to understand the results?
    \item[] Answer: \answerYes{}
    \item[] Justification: The main paper and Appendix~\ref{appendix:dataset} describe the model-specific prompts, source families, and data splits. Intervention locations and evaluation set sizes accompany the corresponding results. Appendix~\ref{appendix:implementation} and the repository documentation describe the evaluation protocol and checkpoint access.

\item {\bf Experiment statistical significance}
    \item[] Question: Does the paper report error bars suitably and correctly defined or other appropriate information about the statistical significance of the experiments?
    \item[] Answer: \answerNo{}
    \item[] Justification: The main causal results are point estimates on fixed held-out sets. We report evaluation sample sizes and norm-matched random controls, but do not report confidence intervals or significance tests for these results.

\item {\bf Experiments compute resources}
    \item[] Question: For each experiment, does the paper provide sufficient information on the computer resources (type of compute workers, memory, time of execution) needed to reproduce the experiments?
    \item[] Answer: \answerYes{}
    \item[] Justification: Appendix~\ref{appendix:implementation:resources} reports approximately 800 NVIDIA B200 GPU-hours for Transcoder training, including debugging and production training. Using the provided checkpoints, the remaining experiments can be reproduced on a single NVIDIA RTX PRO 6000 Blackwell GPU with 96~GB of memory.
    
\item {\bf Code of ethics}
    \item[] Question: Does the research conducted in the paper conform, in every respect, with the NeurIPS Code of Ethics \url{https://neurips.cc/public/EthicsGuidelines}?
    \item[] Answer: \answerYes{}
    \item[] Justification: The research conforms to the NeurIPS Code of Ethics. It uses offline benchmark-based evaluation and causal model interventions, without human-subject studies or collection of sensitive personal data.

\item {\bf Broader impacts}
    \item[] Question: Does the paper discuss both potential positive societal impacts and negative societal impacts of the work performed?
    \item[] Answer: \answerYes{}
    \item[] Justification: Section~\ref{sec:limitations-impacts} discusses potential benefits for understanding and regulating tool use, as well as the possibility of inducing calls inconsistent with user intent.
    
\item {\bf Safeguards}
    \item[] Question: Does the paper describe safeguards that have been put in place for responsible release of data or models that have a high risk for misuse (e.g., pre-trained language models, image generators, or scraped datasets)?
    \item[] Answer: \answerNA{}
    \item[] Justification: The new assets are benchmark-derived paired prompt datasets, analysis code, and Transcoders for interpreting existing models. These research assets do not pose the high-risk release scenarios described in this question.

\item {\bf Licenses for existing assets}
    \item[] Question: Are the creators or original owners of assets (e.g., code, data, models), used in the paper, properly credited and are the license and terms of use explicitly mentioned and properly respected?
    \item[] Answer: \answerYes{}
    \item[] Justification: Original models, benchmarks, and Transcoders are credited in the paper and Appendix~\ref{appendix:implementation:release}, which lists their licenses and source links. Our original release materials use Apache License 2.0; source benchmark content and third-party checkpoints retain their original licenses.

\item {\bf New assets}
    \item[] Question: Are new assets introduced in the paper well documented and is the documentation provided alongside the assets?
    \item[] Answer: \answerYes{}
    \item[] Justification: The paper introduces paired datasets with 500 clean/corrupt pairs for each of the seven primary models. Appendix~\ref{appendix:dataset} documents their sources, splits, prompt formats, and construction. README files, per-model manifests, and the Transcoder model card accompany the assets in the repositories linked in Appendix~\ref{appendix:implementation:release}.

\item {\bf Crowdsourcing and research with human subjects}
    \item[] Question: For crowdsourcing experiments and research with human subjects, does the paper include the full text of instructions given to participants and screenshots, if applicable, as well as details about compensation (if any)? 
    \item[] Answer: \answerNA{}
    \item[] Justification: The paper does not involve crowdsourcing or research with human subjects.

\item {\bf Institutional review board (IRB) approvals or equivalent for research with human subjects}
    \item[] Question: Does the paper describe potential risks incurred by study participants, whether such risks were disclosed to the subjects, and whether Institutional Review Board (IRB) approvals (or an equivalent approval/review based on the requirements of your country or institution) were obtained?
    \item[] Answer: \answerNA{}
    \item[] Justification: The paper does not involve human subjects or participant studies.

\item {\bf Declaration of LLM usage}
    \item[] Question: Does the paper describe the usage of LLMs if it is an important, original, or non-standard component of the core methods in this research? Note that if the LLM is used only for writing, editing, or formatting purposes and does \emph{not} impact the core methodology, scientific rigor, or originality of the research, declaration is not required.
    \item[] Answer: \answerYes{}
    \item[] Justification: Appendix~\ref{appendix:implementation:llm-usage} discloses the use of GPT-5 to generate debugging examples for the data pipeline. The primary mechanistic experiments study seven models from the Qwen3, Qwen3.5, Mistral, and Granite families (Appendix~\ref{appendix:cross-scale}); the additional sparse forward-core replay diagnostic also includes Qwen3-1.7B (Appendix~\ref{appendix:circuit:acdc}).

\end{enumerate}

\end{document}